\documentclass[conference,dvipsnames]{IEEEtran}
\IEEEoverridecommandlockouts

\usepackage{cite} 

\usepackage{amsmath,amssymb,amsfonts} 

\usepackage{algorithmic} 

\usepackage{multirow} 
\usepackage{tabularx} 
\usepackage{booktabs} 
\usepackage{dcolumn} 
\usepackage{colortbl} 

\usepackage{textcomp} 
\usepackage{xcolor} 
\usepackage{bm} 
\usepackage{microtype} 
\usepackage{enumitem} 
\usepackage[shortcuts]{extdash} 

\usepackage{graphicx} 
\usepackage{subcaption} 
\usepackage{circuitikz} 
\usepackage{forest} 
\usetikzlibrary{shapes, positioning} 
\usepackage{svg} 

\usepackage[hidelinks]{hyperref} 

\usepackage{afterpage} 
\usepackage{dblfloatfix} 
\usepackage{balance} 

\usepackage{cleveref} 
\usepackage{orcidlink}

\usepackage{fancyhdr}
\usepackage[left=1.5cm,right=1.5cm,top=2.2cm,bottom=2.5cm]{geometry}

\def\BibTeX{{\rm B\kern-.05em{\sc i\kern-.025em b}\kern-.08em
    T\kern-.1667em\lower.7ex\hbox{E}\kern-.125emX}}

\title{
    SHADE: Deep Density-based Clustering
    \thanks{$^*$First authors with equal contribution in alphabetical order.}
}
\makeatletter
\newcommand{\linebreakand}{%
  \end{@IEEEauthorhalign}
  \hfill\mbox{}\par
  \mbox{}\hfill\begin{@IEEEauthorhalign}
}
\makeatother

\author{
\IEEEauthorblockN{
    Anna Beer$^*$\textsuperscript{1}\orcidlink{0000-0002-6890-997X}, 
    Pascal Weber$^*$\textsuperscript{1,2}\orcidlink{0000-0003-1542-9865},
    \\ 
    Lukas Miklautz\textsuperscript{1}\orcidlink{0000-0002-2585-5895},
    Collin Leiber\textsuperscript{3,4}\orcidlink{0000-0001-5368-5697},
    Walid Durani\textsuperscript{3},
    Christian Böhm\textsuperscript{1}\orcidlink{0000-0002-2237-9969},
    Claudia Plant\textsuperscript{1,5}\orcidlink{0000-0001-5274-8123}
}
\IEEEauthorblockA{
    \textsuperscript{1}\textit{\href{https://dm.cs.univie.ac.at/}{Data Mining and Machine Learning, University of Vienna}}, Vienna, Austria\\
    \textsuperscript{2}\textit{\href{https://docs.univie.ac.at/}{UniVie Doctoral School Computer Science}}, Vienna, Austria\\
    \textsuperscript{3}\textit{\href{https://www.dbs.ifi.lmu.de/cms/}{Database Systems and Data Mining, LMU Munich}}, Munich, Germany\\
    \textsuperscript{4}\textit{\href{https://mcml.ai/}{Munich Center for Machine Learning}}, Munich, Germany\\
    \textsuperscript{5}\textit{\href{https://datascience.univie.ac.at/}{ds:UniVie}}, Vienna, Austria\\
    \href{mailto:anna.beer@univie.ac.at,pascal.weber@univie.ac.at?subject=[SHADE]\%20I\%20want\%20to\%20reach\%20out\&cc=claudia.plant@univie.ac.at;christian.boehm@univie.ac.at}{\{firstname.lastname\}@univie.ac.at}, \href{mailto:leiber@dbs.ifi.lmu.de,durani@dbs.ifi.lmu.de?subject=[SHADE]\%20I\%20want\%20to\%20reach\%20out\&cc=anna.beer@univie.ac.at;pascal.weber@univie.ac.at;claudia.plant@univie.ac.at;christian.boehm@univie.ac.at}{\{lastname\}@dbs.ifi.lmu.de}
}
}

\begin{document}

\maketitle
\thispagestyle{plain} 

\rhead{Beer, Weber, et al.}
\lhead{SHADE: Deep Density-based Clustering}
\pagestyle{fancy}

\begin{abstract}
Detecting arbitrarily shaped clusters in high-dimensional noisy data is challenging for current clustering methods. We introduce SHADE (Structure-preserving High-dimensional Analysis with Density-based Exploration), the first deep clustering algorithm that incorporates density-connectivity into its loss function. Similar to existing deep clustering algorithms, SHADE supports high-dimensional and large data sets with the expressive power of a deep autoencoder. In contrast to most existing deep clustering methods that rely on a centroid-based clustering objective, SHADE incorporates a novel loss function that captures density-connectivity. SHADE thereby learns a representation that enhances the separation of density-connected clusters.
SHADE detects a stable clustering and noise points fully automatically without any user input. It outperforms existing methods in clustering quality, especially on data that contain non-Gaussian clusters, such as video data. Moreover, the embedded space of SHADE is suitable for visualization and interpretation of the clustering results as the individual shapes of the clusters are preserved.
\end{abstract}


\section{Introduction}

\begin{figure}[ht]

    \def\leftspace{1.2cm}
    \def\bottomspace{0.95cm}
    \def\rightspace{0.2cm}
    \def\topspaceF{0cm}
    \def\topspace{-1.2cm}

    \vspace{-0.1cm}
    \centering

    \begin{subfigure}[b]{0.32\columnwidth}
    \centering
    \includegraphics[width=1\columnwidth, trim={3.8cm 1.9cm 3.2cm 2cm}, clip]{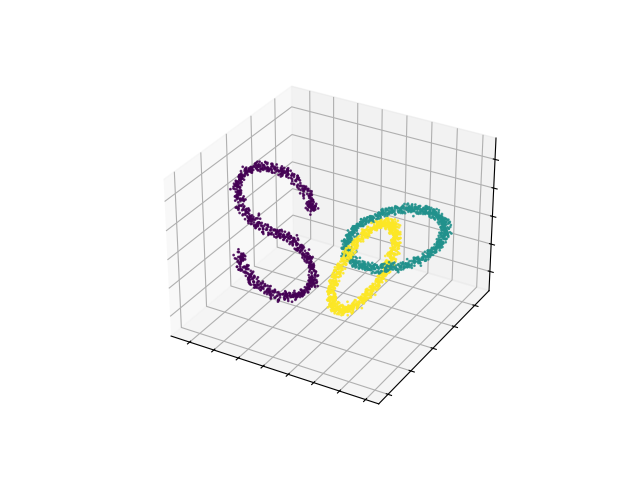}
    \caption{3d dataset}
    \end{subfigure}
    \begin{subfigure}[b]{0.32\columnwidth}
    \includegraphics[
        width=\columnwidth, 
        trim={\leftspace{} \bottomspace{} \rightspace{} \topspaceF{}}, 
        clip,
    ]{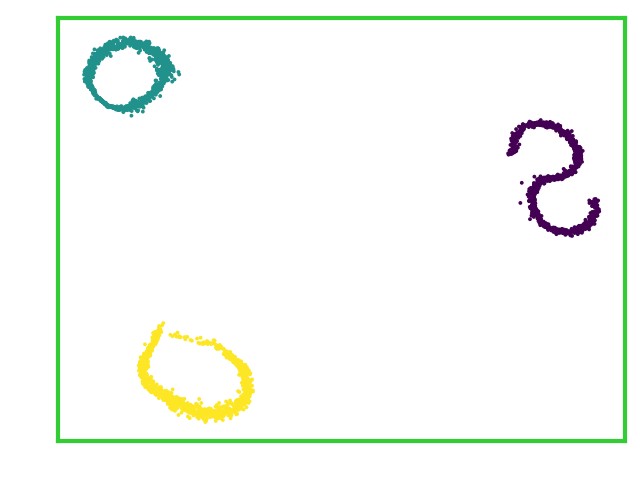}
    \caption{SHADE ($1.00$)}
    \end{subfigure}
    \begin{subfigure}[b]{0.32\columnwidth}
    \includegraphics[
        width=\columnwidth, 
        trim={\leftspace{} \bottomspace{} \rightspace{} \topspaceF{}}, 
        clip,
    ]{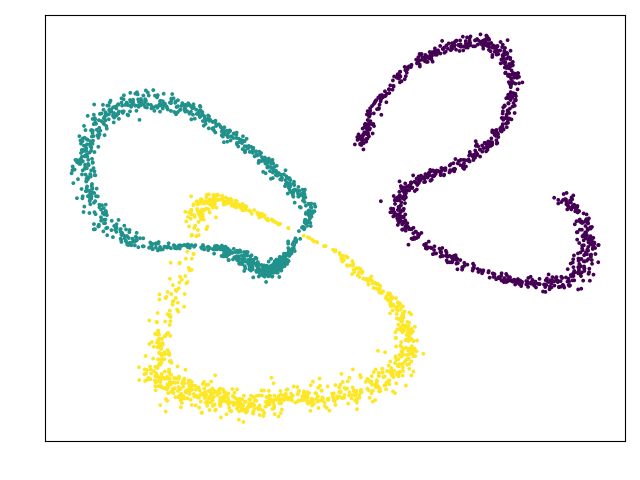}
    \caption{Autoencoder}
    \end{subfigure}
    
    \begin{subfigure}[c]{0.32\columnwidth}
    \includegraphics[
        width=\columnwidth, 
        trim={\leftspace{} \bottomspace{} \rightspace{} \topspace{}}, 
        clip,
    ]{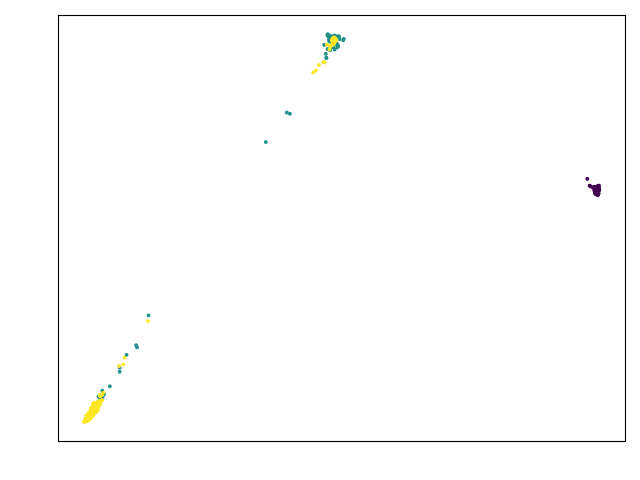}
    \caption{DEC ($0.59$)}
    \end{subfigure}
    \begin{subfigure}[c]{0.32\columnwidth}
    \includegraphics[
        width=\columnwidth, 
        trim={\leftspace{} \bottomspace{} \rightspace{} \topspace{}}, 
        clip,
    ]{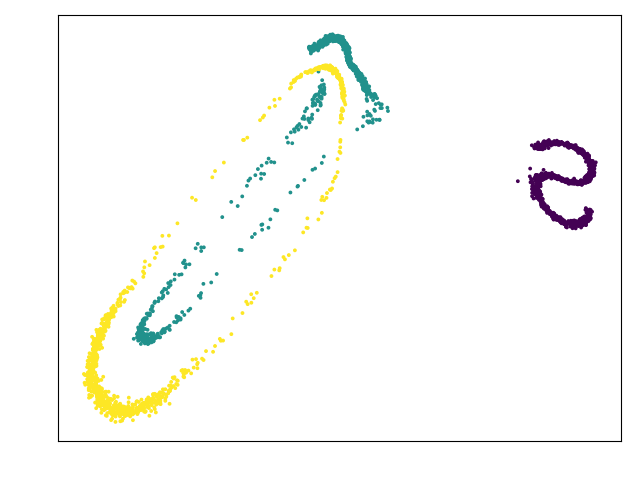}
    \caption{IDEC ($0.64$)}
    \end{subfigure}
    \begin{subfigure}[c]{0.32\columnwidth}
    \includegraphics[
        width=\columnwidth, 
        trim={\leftspace{} \bottomspace{} \rightspace{} \topspace{}}, 
        clip,
    ]{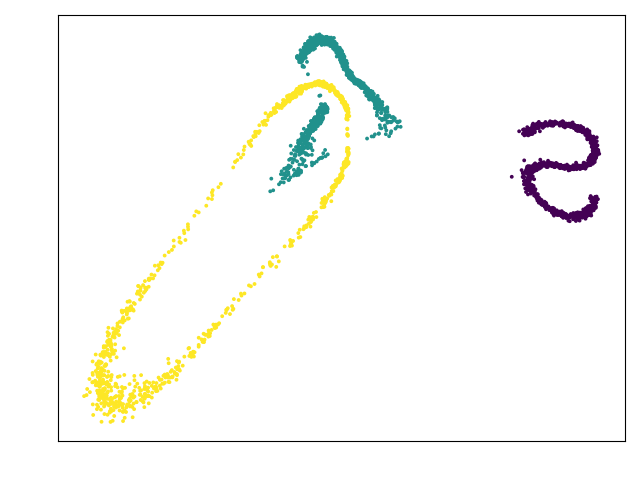}
    \caption{DCN ($0.64$)}
    \end{subfigure}

    \begin{subfigure}[c]{0.32\columnwidth}
    \includegraphics[
        width=\columnwidth, 
        trim={\leftspace{} \bottomspace{} \rightspace{} \topspace{}}, 
        clip,
    ]{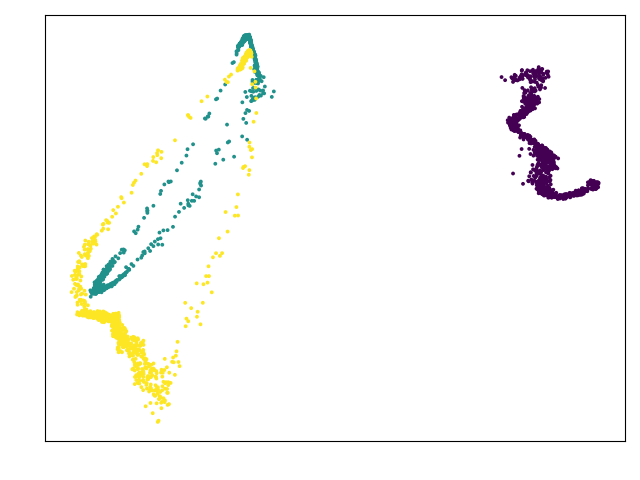}
    \caption{DipEncoder ($0.60$)}
    \end{subfigure}
    \begin{subfigure}[c]{0.32\columnwidth}
    \includegraphics[
        width=\columnwidth, 
        trim={\leftspace{} \bottomspace{} \rightspace{} \topspace{}}, 
        clip,
    ]{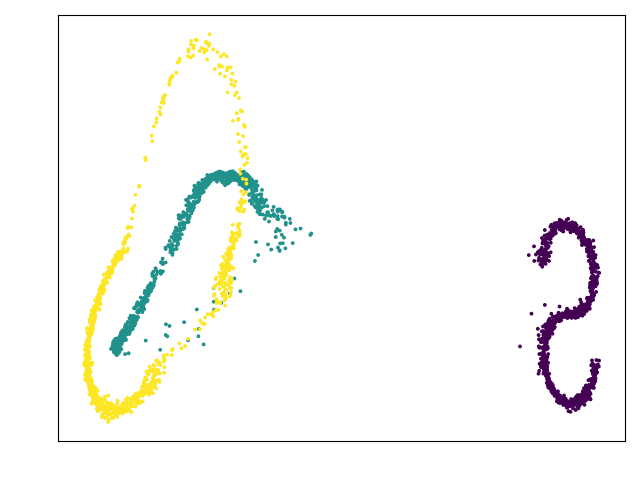}
    \caption{DipDECK ($0.57$)}
    \end{subfigure}
    \begin{subfigure}[c]{0.32\columnwidth}
    \includegraphics[
        width=\columnwidth, 
        trim={\leftspace{} \bottomspace{} \rightspace{} \topspace{}}, 
        clip,
    ]{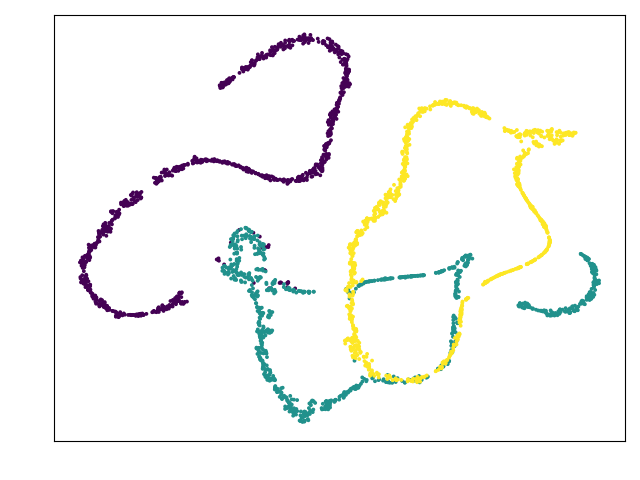}
    \caption{DDC ($0.25$)}
    \end{subfigure}
    \caption{3d dataset (a) and its 2d embedding created by our algorithm SHADE (b), a regular autoencoder (c), and its competitors (d)-(i); colors imply ground truth clusters and numbers in brackets show the clustering quality measured by ARI. SHADE separates the clusters and keeps their shapes, whereas other methods merge them or change the shape entirely.}
    \label{fig:motivation}
\end{figure}

\begin{figure*}[ht]

    \def\leftspace{0.2cm}
    \def\bottomspace{0.45cm}
    \def\rightspace{0.2cm}
    \def\topspaceF{0cm}
    \def\topspace{-0.8cm}

    \centering

    \begin{subfigure}{0.325\linewidth}
    \centering
     \includegraphics[
        width=\columnwidth, 
        trim={\leftspace{} \bottomspace{} \rightspace{} \topspace{}}, 
        clip,
    ]{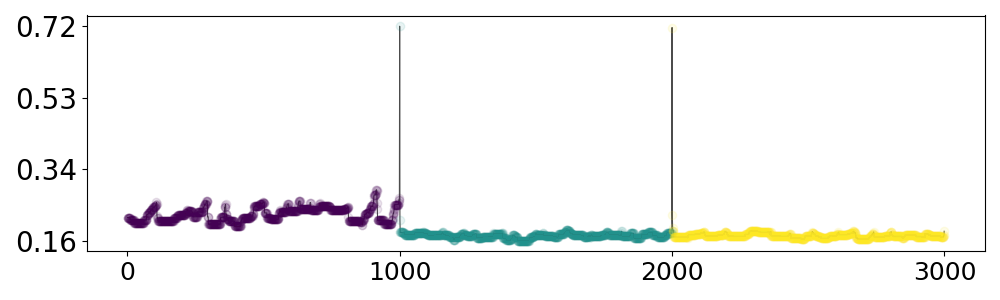}
    \caption{3d dataset (0.98)}
    \end{subfigure}
    \begin{subfigure}{0.325\linewidth}
     \includegraphics[
        width=\columnwidth, 
        trim={\leftspace{} \bottomspace{} \rightspace{} \topspace{}}, 
        clip,
    ]{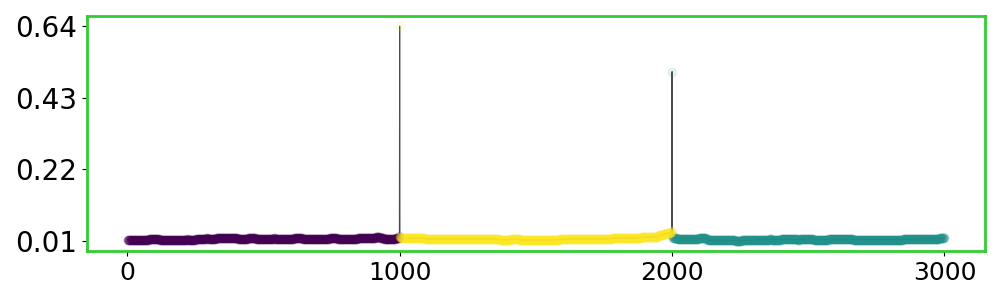}
    \caption{SHADE (1.00)}
    \end{subfigure}
    \begin{subfigure}{0.325\linewidth}
     \includegraphics[
        width=\columnwidth, 
        trim={\leftspace{} \bottomspace{} \rightspace{} \topspace{}}, 
        clip,
    ]{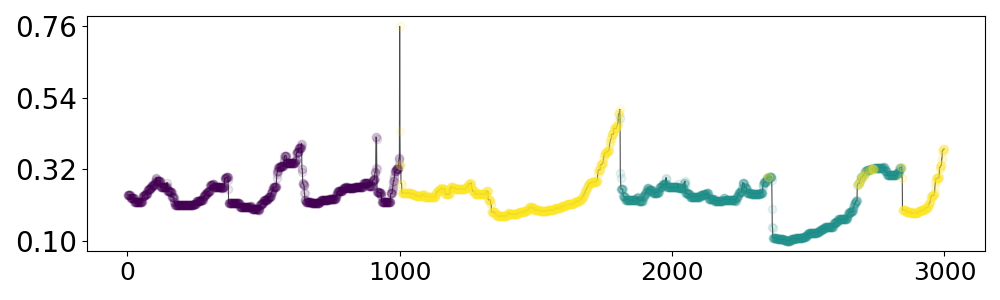}
    \caption{Autoencoder (0.58)}
    \end{subfigure}
    
    \begin{subfigure}{0.325\linewidth}
     \includegraphics[
        width=\columnwidth, 
        trim={\leftspace{} \bottomspace{} \rightspace{} \topspace{}}, 
        clip,
    ]{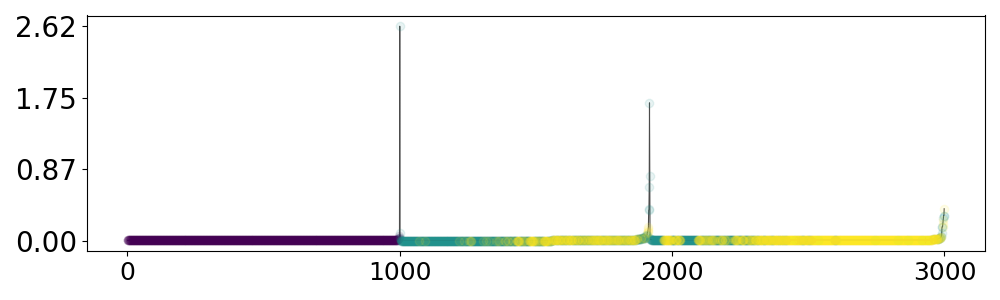}
    \caption{DEC (0.26)}
    \end{subfigure}
    \begin{subfigure}{0.325\linewidth}
     \includegraphics[
        width=\columnwidth, 
        trim={\leftspace{} \bottomspace{} \rightspace{} \topspace{}}, 
        clip,
    ]{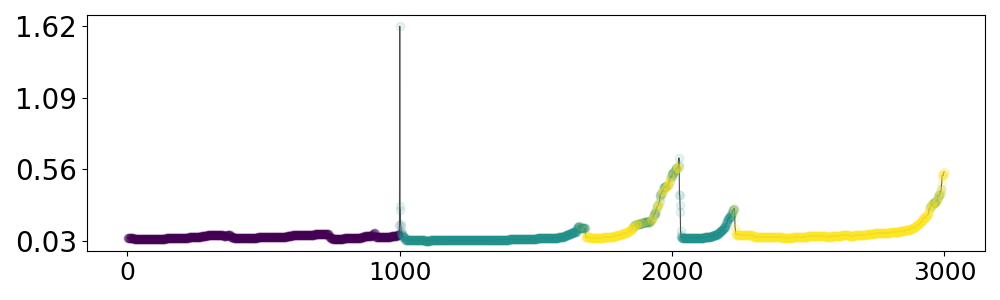}
    \caption{IDEC (0.26)}
    \end{subfigure}
    \begin{subfigure}{0.325\linewidth}
     \includegraphics[
        width=\columnwidth, 
        trim={\leftspace{} \bottomspace{} \rightspace{} \topspace{}}, 
        clip,
    ]{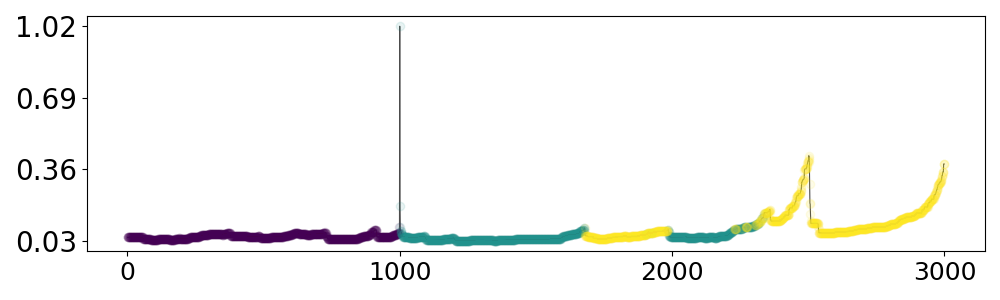}
    \caption{DCN (0.58)}
    \end{subfigure}
    
    \begin{subfigure}{0.325\linewidth}
     \includegraphics[
        width=\columnwidth, 
        trim={\leftspace{} \bottomspace{} \rightspace{} \topspace{}}, 
        clip,
    ]{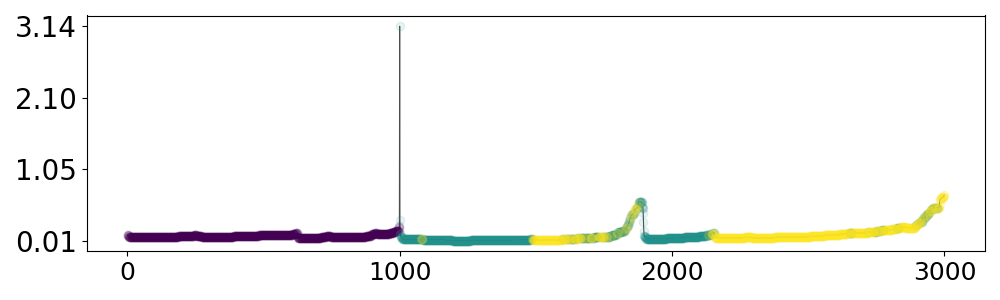}
    \caption{DipEncoder (0.41)}
    \end{subfigure}
    \begin{subfigure}{0.325\linewidth}
     \includegraphics[
        width=\columnwidth, 
        trim={\leftspace{} \bottomspace{} \rightspace{} \topspace{}}, 
        clip,
    ]{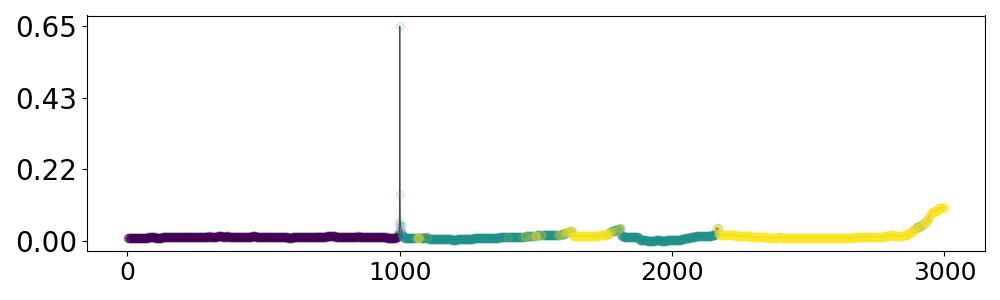}
    \caption{DipDECK (0.67)}
    \end{subfigure}
    \begin{subfigure}{0.325\linewidth}
     \includegraphics[
        width=\columnwidth, 
        trim={\leftspace{} \bottomspace{} \rightspace{} \topspace{}}, 
        clip,
    ]{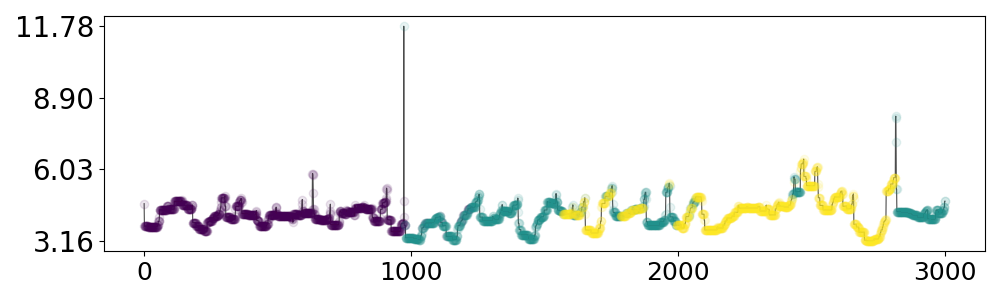}
    \caption{DDC (0.44)}
    \end{subfigure}

    \caption{OPTICS~\cite{optics} reachability plots of the 3d dataset (a) and of its 2d embedding created by our algorithm SHADE (b), a regular autoencoder (c), and our competitors (d)-(i); colors imply ground truth clusters and numbers in brackets show the cluster separability measured by the density cluster separability index (DCSI). SHADE retains the original 3d dataset's overall reachability structure and enhances cluster separability. In contrast, other methods merge the two intertwined clusters or fail to separate the intertwined clusters correctly. Additionally, the other methods reduce the overall cluster separability.
    }
    \label{fig:reachability_plots}
\end{figure*}

Density-based clustering considers clusters as areas of high object density that are separated by areas of low object density; see, for example, the 3d dataset in Figure \ref{fig:motivation}(a) with two intertwined rings and one s-shaped cluster. While this is a synthetic example, density-based clusters are also common in real-world data: e.g., clusters following geographical structures or evolutionary development processes \cite{DBLP:journals/widm/CampelloKSZ20}. 
Real-world data typically exhibits a much higher dimensionality than our synthetic example, often in the order of hundreds or thousands of dimensions. Density-based structures are especially hard to find in such high-dimensional spaces because of the empty-space problem~\cite{steinbach2004challenges}.

We introduce SHADE (\textbf{S}tructure-preserving \textbf{H}igh-dimensional \textbf{A}nalysis with \textbf{D}ensity-based \textbf{E}xploration), a novel clustering method that integrates a density-connectivity loss into the training of a deep autoencoder. Many recent deep clustering methods, such as \cite{DEC, IDEC, dipdeck}, exploit the expressive power of deep autoencoders to learn cluster-friendly representations of high-dimensional data. SHADE is the first method that focuses on learning a representation that enhances density-based clusters. Figure \ref{fig:motivation}(b) shows the representation learned by SHADE when reducing the dimensionality of our 3d synthetic dataset to 2d. SHADE enhances the density-based cluster structure by improving the separation of the three clusters while preserving their individual shapes. Using the density-connectivity information that is part of SHADE's loss function, our algorithm also outputs a density-based clustering, which is perfect on this dataset with an ARI of~$1.0$. 

Note that real-world data often has more dimensions than traditional density-based methods as, e.g., DBSCAN~\cite{dbscan} and OPTICS~\cite{optics}, can handle -- however, they work well on our 3d toy dataset.
Figure \ref{fig:reachability_plots}(a) shows the OPTICS reachability plot of the 3d data. The colors correspond to the ground truth classes, see also Figure \ref{fig:motivation}.  All three clusters are visible as class-pure valleys in the plot that are separated by high reachability distances. To add a quantitative measure, we also report the DCSI~\cite{dcsi}, which is an internal validity measure for density-based clustering that scales between $0$ and~$1$. DCSI measures the density-connectedness within classes and the separation between classes. 
In the original 3d space, the ground truth clusters are well density-separated, expressed by a very high DCSI of $0.98$.

Figure \ref{fig:reachability_plots}(b) shows that SHADE improves the density-connectivity within the classes, which results in a perfect DCSI measure of $1.0$. At the same time, SHADE preserves the shapes of the original clusters, see Figure \ref{fig:motivation}(b). 

Why not just use an autoencoder for representation learning followed by density-based clustering? 
Autoencoders minimize the reconstruction error of individual data points using some objective function. However, in clustering, we want to group similar points, which is a different objective. In the 2d embedding of the autoencoder in \Cref{fig:motivation}(c), the two rings are connected, leading to a rather low DCSI of $0.58$. The corresponding OPTICS plot in Figure \ref{fig:reachability_plots}(c) highlights this problem: some of the yellow points are in the same valley as some of the green points, i.e., they have a closer connection to points of the wrong ground truth cluster. Thus, the 2d embedding of the data by a classic autoencoder does not allow a correct, density-based clustering anymore.

Therefore, recent research papers on deep clustering have integrated specific clustering losses into the objective function of autoencoders. The vast majority of approaches rely on a centroid-based cluster notion, similar to $k$-Means, e.g., \cite{DEC, IDEC, DCN, DKM, Acedec}. Figures \ref{fig:motivation}(d)-(f) and \ref{fig:reachability_plots}(d)-(f) demonstrate that these methods tend to fail on data with density-based clusters. In the embedded space of DEC \cite{DEC}, the clusters seem to be well separated, but the shapes are completely destroyed by forcing them to fit Gaussian distributions. The ARI measure of $0.59$ and the OPTICS plot in the embedded space demonstrate that DEC cannot separate the two rings. The corresponding reachability plot has two valleys, but these clusters consist of a mixture of points from both classes. Consequently, the DCSI measure is very low, with a value of only $0.26$. IDEC and DCN do not enforce Gaussianity as strictly as DEC but also employ centroid-based losses. Both methods preserve the 's' shape but fail to separate the rings.

The recently proposed methods DipDECK \cite{dipdeck} and DipEncoder \cite{dipencoder} support a more flexible cluster model by relying only on modalities instead of specific distributions. DipDECK further offers the benefit that users do not need to select the number of clusters as an input parameter. However, both methods perform similarly to DEC and IDEC in terms of ARI and DCSI, cf. Figures \ref{fig:motivation}(g), \ref{fig:motivation}(h), \ref{fig:reachability_plots}(g), and \ref{fig:reachability_plots}(h). In the embedded space of the autoencoder, both rings heavily overlap. Therefore, considering modalities does not help. 

In Figures \ref{fig:motivation}(i) and \ref{fig:reachability_plots}(i), we see the result of DDC \cite{DDC_deepDensity}, a recent sequential approach that is more elaborated than just using density-based clustering on the autoencoder embedding. After autoencoder pretraining, t-SNE is applied to enhance cluster structures and to reduce the dimensionality. As its last step, DDC clusters the data in the latent space with a variation of the DensityPeaks algorithm. On our example dataset, this approach cannot separate the intertwined rings either. The initial problems emerge from the mismatch of the objective function of the autoencoder and the goal of density-based clustering, which cannot be cured by t-SNE, resulting in an ARI of $0.25$ and a DCSI of $0.44$.

Therefore, we propose SHADE, the first approach combining the benefits of density-based and deep clustering. 
Our main contributions are as follows: 
\begin{enumerate}[leftmargin=*]
    \item SHADE is the first deep density-based clustering algorithm, enabling joint representation learning and density-based clustering.
    \item The representation learned by SHADE enhances the separation of density-based clusters while preserving their density-connectivity.
    \item Empowered by a deep autoencoder, SHADE supports density-based clustering of high-dimensional data such as images and videos.
    \item Inherited from density-based clustering, SHADE naturally supports noise.
    \item SHADE detects density-based clusters and noise without requiring the user to know the number of clusters or percentage of noise beforehand
\end{enumerate}

\section{A novel deep density-based clustering method}\label{sec:SHADEmain}
Aiming to find clusters that consist of a series of similar objects (instead of objects that are \textit{all} similar to only \textit{one} centroid) makes fewer assumptions about the data: for a cluster assignment, a point just needs to be close to a few of \textit{any} other points in the cluster.
The shapes of clusters are less restricted, and thus, density-based methods can detect important structures that are not detectable with centroid-based methods. 
However, capturing density-connected clusters in high-dimensional space is hard and can be slow. 
Hence, we first learn a low-dimensional embedding where we capture and preserve the density-connected structures within the high-dimensional space. Afterward, we cluster the embedded data points in the low-dimensional space.

\begin{figure}[t]
    \centering
    \includegraphics[trim= 6 5 8 6 , clip , width=0.99\linewidth]{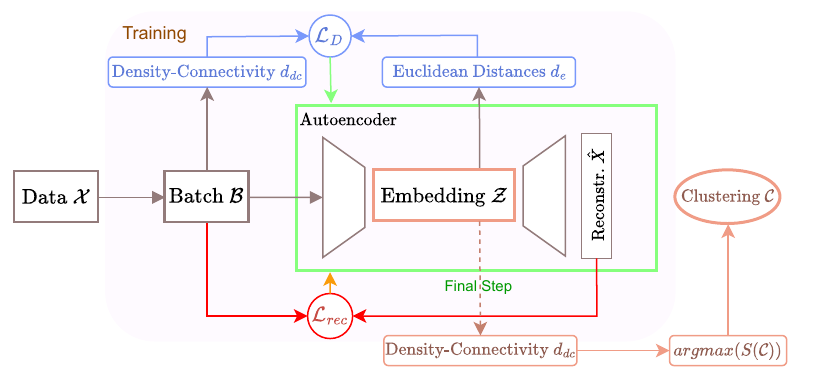}
    \caption{
    SHADE optimizes the loss functions $\mathcal{L}_D$ and $\mathcal{L}_{rec}$ simultaneously in a batch-wise manner. $\mathcal{L}_D$ aligns the density-connectivity in the original space with the Euclidean distances in the embedding. 
    $\mathcal{L}_{rec}$, on the other hand, enforces that the original high-dimensional spatial structure or shape of the clusters is preserved in the learned embedding, allowing an accurate reconstruction.
    The final clustering $\mathcal{C}$ is obtained by selecting the clustering with the highest stability $S(\mathcal{C})$ based on the density-connectivity metric $d_{dc}$.
    }
    \label{fig:overview}
\end{figure}

SHADE's main steps are shown in \Cref{fig:overview}, which we explain in more detail throughout this section.

\subsection{Challenges of Density-Connected Structures}\label{sec:challenge}
We identify three main challenges that come with density-connected structures in high-dimensional data:

\smallskip

\subsubsection{Separability between intra- and inter-cluster distances} For density-connected clusters, the (pairwise) Euclidean distance between points does not necessarily correlate with cluster membership:
Clusters of arbitrary shape can be elongated, yielding large Euclidean intra-cluster distances that are potentially higher than inter-cluster distances.
E.g., the Euclidean distance between points within a circular cluster as in \Cref{fig:motivation} can be as large as the diameter of the circle.
However, the Euclidean distance between the closest points of different clusters is only half the diameter.

\subsubsection{Preserving distances between structurally relevant points} 
Especially for points that lie between two clusters, slight offsets in the embedding can have grave consequences regarding the density-connected cluster structure. 
In contrast, points at the border of the data space might not change the clustering result at all.
Thus, the preservation of distances between points that are crucial for the clustering structure should be prioritized.

\subsubsection{Non-contractible and intertwined clusters}
Density-connected clusters might consist of intertwined topological structures that cannot be separated by AEs that produce a smooth transition of the data to the embedding.
AEs based on the reconstruction loss are powerful tools for clustering if the points of each cluster lie in a \textit{contractible} space~\cite{topology}. 
Intuitively, a topological space is contractible if it can be shrunk to a point by a continuous function. A smooth AE could find such a function or embedding when given an according loss function.  
After contracting each cluster's space to a point within that space, i.e., embedding the points, the clusters are trivially separable in the embedding. 
(e.g., centroid-based deep clustering approaches might map all points of a cluster onto its center). 
Density-connected clusters imply \textit{simply connected} spaces that might be non-contractible, e.g., an annulus or a ring, as shown in \Cref{fig:motivation}.
These clusters cannot be contracted onto a point, and if they are intertwined as in \Cref{fig:motivation}, they are not easily separable in any embedding given by a smooth function.

\subsection{Capturing Density-Connectivity}\label{sec:captureDC}
To preserve and enhance density-connected structures and tackle the aforementioned challenges, we introduce a novel density-connectivity loss $\mathcal{L}_d$ in \Cref{sec:DCLoss}.
This loss needs to capture the density-connectivity and enable the AE to transfer the relevant information from the high-dimensional space into the low-dimensional embedding. 
To define it, we need some theoretical background knowledge on density-connectivity (\Cref{sec:back-dc} and \Cref{sec:backHDB}).

\subsubsection{Background: Classic Density-Connectivity}\label{sec:back-dc}
Density-connected clusters are defined as maximal subsets of connected \textit{core points}~\cite{dbscan}. For parameters $\varepsilon \in \mathbb{R}_{>0}$ and $\mu \in \mathbb{N}$ that are usually given by users\footnote{$\mu$ is also known as $minPts$ in some literature}, core points are defined as points with at least $\mu$ points in their $\varepsilon$-range. 
The core distance of a point $x \in \mathcal{X}$, $core\_dist(x)$ gives the distance to its $\mu$-th nearest neighbor. 
Core points are \textit{connected} if they are closer than $\varepsilon$. 
The transitive hull of density-connected core points defines a density-connected cluster and contains all points that are \textit{density-reachable}, i.e., that have a `chain' of core points connecting them. 
These concepts are used in a variety of literature, e.g., \cite{dbscan, dbscan-rev-rev, anydbc, schubert_dbscanMFSC_2018}.
The mutual reachability distance $d_m$ is often used in methods building on top of it (e.g., \cite{optics, hdbscan, hdbscan_long, schubert_dbscanMFSC_2018}) and is defined as $d_m(x,y)=\max(d_{eucl}(x,y), core\_dist(x), core\_dist(y))$ for $x,y \in \mathcal{X}$, where $d_{eucl}$ is the Euclidean distance.

\subsubsection{Background: Hierarchical Density-Connected Structures}\label{sec:backHDB}
Analogous to hierarchical single-linkage clustering~\cite{murtagh1983survey_hierarchical}, the minimum spanning tree (MST) captures the hierarchical structure of density-connected clusters. While single-linkage clustering uses the Euclidean distance graph as a basis for the MST, for density-connected structures, the hierarchy is based on the MST of the graph given by the mutual reachability distance \cite{dcdist, wishart_density_SL}.

The minimax path distances on this MST imply a tree metric $d_{dc}$ that captures the smallest $\varepsilon$ s.t. two points are density-reachable~\cite{dcdist}. 
We can store the values of $d_{dc}$ in a tree $\mathcal{T}_d$, as $d_{dc}$ is a tree-metric. Its root node contains the length of the largest edge in the MST, and its leaves represent the points in the dataset. The distance $d_{dc}(x,y)$ between two points is stored in their lowest common ancestor in $\mathcal{T}_d$, and without loss of generality, we can assume $\mathcal{T}_d$ to be a binary tree.
Capturing the density-connectivity via $\mathcal{T}_d$ has several advantages: 
It can leverage some problems connected to the curse of dimensionality, e.g., after constructing the MST, all subsequent computations regarding the density are independent of the original dimensionality.
Furthermore, $\mathcal{T}_d$ captures the full hierarchical structure of the data instead of a single partitioning, allowing it to capture noise and clusters of different densities. 

\subsection{Density-Connectivity Loss}\label{sec:DCLoss}
We propose a novel loss function that captures the relevant information of density-connected structures in the high-dimensional space and transfers them into the embedding. In the following, we describe the key ideas behind our proposed density-connectivity loss $\mathcal{L}_d$ (Equation \ref{eq:DC_Loss_orig}) that is able to tackle all these challenges.

The $d_{dc}$ metric captures density-connectivity, with inter-cluster distances consistently larger than intra-cluster distances~\cite{dcdist}, a promising solution for \textbf{challenge (1)}. 
However, using this property effectively for the loss function is not trivial. As $d_{dc}$ is an ultrametric, any data $\mathcal{X}$ can be trivially embedded into the one-dimensional space with full preservation of all pairwise $d_{dc}$. Thus, preserving the $d_{dc}$ from the original space to the embedding would not yield meaningful structures. 
Thus, we use $d_{dc}$ only in the original space while using the Euclidean distance in the embedded space. This allows learning an embedding where points within a density-connected structure are close, and points in different clusters are far apart. 

To address the varying relevance of points, \textbf{challenge~(2)}, note that only distances along the MST of the mutual reachability distance are important for the hierarchy of high-dimensional density-connected structures.
Employing $d_{dc}$ in the original space automatically omits irrelevant distance values between points that are not directly connected in the MST. Thus, only the distances relevant to the (hierarchical) clustering structure are explicitly used by the tree metric. 
Therefore, the relevant points' influence is captured in $d_{dc}$. 

\textbf{Challenge (3)}, the problem of non-contractible and intertwined clusters, is solved by reducing the fully connected distance graph to the MST:
Capturing the density-connected structure as a path-based distance hides the information about the entanglement of structures and reduces it to information about their connectivity. 
E.g., regard the two intertwined rings from  \Cref{fig:motivation}, where both circles have the same radius $r$, which also corresponds to the distance of the points of one ring to its center. 
The only information stored about the relationship between the two rings is the length of the link in the MST between them. Thus, any point of one ring has exactly the same $d_{dc}=r$ to any point of the other ring.
Note that a dataset with two rings that are \textit{not} intertwined and have an Euclidean distance of $3r$ between their centers has the exact same distance matrix regarding $d_{dc}$.
Therefore, relying on the MST allows the autoencoder to split apart the rings and similar intertwined structures.
Thus, $d_{dc}$ carries all the information that we need for successfully tackling challenges (1) to (3), and we get an embedding with the desired properties by minimizing the following density-connectivity loss:
\begin{equation}\label{eq:DC_Loss_orig}
\mathcal{L}_d = \frac{1}{\lvert \mathcal{X} \rvert^2} \sum_{x_i,x_j \in \mathcal{X}}  \left( \  d_{dc}(x_i,x_j)-d_{eucl}(z_i,z_j) \ \right)^2
\end{equation}  

for any points $x_i,x_j$ in $\mathcal{X}$ and their corresponding embedded points $z_i=enc(x_i),z_j=enc(x_j)$. 

However, using the whole dataset at once is often not feasible when training the loss function. Hence, it is optimized in a batch-wise manner. We can then approximate our original loss functions with the following loss:

\begin{equation}\label{eq:DC_Loss}
\mathcal{L}_d = \frac{1}{\lvert \mathcal{B} \rvert^2} \sum_{x_i,x_j \in \mathcal{B}}  \left( \  d_{dc}(x_i,x_j)-d_{eucl}(z_i,z_j) \ \right)^2
\end{equation}  
for any points $x_i,x_j$ in the batch $\mathcal{B}$ and their corresponding embedded points $z_i=enc(x_i),z_j=enc(x_j)$. 
In \Cref{sec:ablation_batch_size}, we show that this approach yields stable results across various batch sizes, making it valuable for real-world applications.

\subsection{Preserving the Structure in the Embedding}
By minimizing the density-connectivity loss, we preserve the inter-cluster distances and compress the points within one cluster together such that the distance between them corresponds to the $d_{dc}$ distance in the high-dimensional space. That way, we also achieve better separability between clusters. 

To preserve the overall cluster structure in the low-dimensional embedding, e.g., their shapes, we additionally minimize the reconstruction error: 
\begin{equation}\label{eq:L_rec}
\mathcal{L}_{rec} = \frac{1}{\lvert \mathcal{B} \rvert} \sum_{x_i \in \mathcal{B}}   \left\lVert x_i - \hat{x}_i \right\rVert_2^2
\end{equation}
where $\hat{x_i}$ = $dec(enc(x_i))$ is the reconstruction of $x_i$. 

As shown in \Cref{fig:motivation}, the AE, which only minimizes the reconstruction loss $\mathcal{L}_{rec}$, already retains the cluster structure to a certain degree but fails to separate the intertwined clusters. We can enforce the separation of the different density-connected clusters and preserve the cluster structure by including our novel density-connectivity loss $\mathcal{L}_{d}$. 

\smallskip

Training on this combined loss function creates an embedding that preserves and enhances density-connected structures. 
This embedding is very suitable for visualizing data, as it preserves the shapes and cluster structure.
It is, furthermore, the foundation for finding density-connected structures, which we do by applying a novel clustering technique described in \Cref{sec:fullyAutomaticClustering} in the embedding. 
This technique unites multiple desirable and hard-to-reach properties: it returns a stable clustering, it detects noise, it finds the number of clusters fully automatically, and it allows clusters with different densities. 
Combining deep learning with an incorporated density-connectivity loss and this technique, we are able to cluster very high-dimensional data by enhancing the density-connectivity and the shape of clusters in the embedding space.

\subsection{Fully Automatic Clustering and Noise Detection in the Embedded Space}\label{sec:fullyAutomaticClustering}
We can now perform the final clustering step in the embedded space obtained with the steps described above.
Note that we cannot simply apply $k$-Means in the embedded space like some of our competitors, as the density-connected structures from the high-dimensional space are preserved. As a result, clusters are typically not convex, rendering centroid-based clustering methods ineffective.
Instead, we detect the density-based structures in the low-dimensional embedding based on the cluster hierarchy given by $\mathcal{T}_d$ (see \Cref{sec:backHDB}).
Note that $\mathcal{T}_d$ offers a variety of possible clusterings, some of which are better than others. 
For example, DBSCAN-like clustering results in the embedding can be obtained by thresholding the tree at a user-given $\varepsilon$.
However, this approach prevents the detection of clusters with different densities, and $\varepsilon$ is hard to choose.

Thus, automatically detecting the best partitioning for a given tree is desirable. 
We impose two additional challenges that are especially important for exploratory data analysis, as it is typical for our unsupervised setting: 1) the number of clusters should be detected automatically, and 2) points that do not belong to a cluster should not be assigned to a cluster, i.e., the algorithm needs to be able to detect noise.
Note that to the best of our knowledge, no other deep clustering algorithm fulfills these requirements. Our novel method, SHADE, is the first deep clustering method that inherently detects noise and is simultaneously part of a small group of deep clustering algorithms that can automatically identify the number of clusters.
SHADE solves both problems by introducing a novel measure for stability and a method to find the most \textit{stable} clustering.

\subsubsection{Stability of Clusters}\label{sec:stabilityOfClusters}
For SHADE, we define \textit{stable} clusters as clusters that persist for a large variety of densities. 
Our stability value corresponds to the range of densities that produce the same cluster besides bordering or noise points.
We compute it by first reducing the cluster hierarchy to nodes relevant to the structure of the tree and then regard the values of nodes in consecutive levels in this simplified \textit{structure tree} as explained in the following.
For other notions of stability and delimitation to those, we refer to \Cref{sec:rel:stability}.

\textbf{Notation:} 
A group of leaves $l(a)$ is defined by its lowest common ancestor node $a$ in a tree $\mathcal{T}$ and has $|l(a)|$ many members. The node $a$ stores the distance value $\mathcal{T}(a)$.
We refer to the parent node of $a$ with $p(a)$ and the children with $ch(a)$.
The nodes $l(a) \in \mathcal{T}_d$ build a proper\footnote{According to the definitions of DBSCAN$^{(*)}$~\cite{dbscan}} density-connected cluster for all $\varepsilon \geq \mathcal{T}_d(a)$. 
\paragraph{Structure Tree}
To capture the relevant cluster splits (rather than splits between a cluster and bordering noise), we define a structure tree $\mathcal{T}$. It exclusively contains nodes of $\mathcal{T}_d$ where both children have at least $\mu$ leaves.
$\mathcal{T} = \{a \in \mathcal{T}_d |~ \forall ch \in ch(a): \lvert l(ch) \rvert  \geq \mu \}$.
Nodes $b \in \mathcal{T}_d$ that do not fulfill this property are skipped in $\mathcal{T}$, and if they have children, their leaves $l(b)$ are assigned to the topmost child of $b$ that fulfills the property. 
This assigns bordering noise points within a distance $< \mathcal{T}_d(p(a))$ to their closest cluster. 
If, after this step, a node only has one remaining child, we merge the child node with its parent and keep the $d_{dc}$ value of the child node. 

\paragraph{Stability}
The range of densities for which points~$\in~l(a)$ build a density-connected cluster $c$ is the absolute difference between $\mathcal{T}_d(a)^{-1}$ and $ \mathcal{T}_d(p(a))^{-1}$, considering the density given by $1 / \varepsilon$. 
Regarding the structure tree $\mathcal{T}$ and including the number of points in a cluster $c$ gives us the stability $S(c)$\footnote{Note the ultrametric property of $\mathcal{T}_d$ which determines that $\mathcal{T}_d(p(a)) \geq \mathcal{T}_d(a)$ for any $a$ (besides the root node).}:
\begin{equation}\label{eq:stability}
    S(c)= \left( \frac{1}{\mathcal{T}_d(a)}- \frac{1}{\mathcal{T}_d(p(a))} \right) \cdot \lvert l(a) \lvert 
\end{equation}
The stability of a clustering $\mathcal{C}$ is the sum of its cluster's stabilities:
$S(\mathcal{C})=\sum_{c \in C} S(c)$

\subsubsection{Automatically finding most stable clustering}
Finding the most stable clustering efficiently is not trivial, as there are many potential clusterings for a given hierarchy. 
Note that we look for a flat clustering, where each cluster is defined by a root node containing all descendant leaves of the root node. 
Furthermore, especially for exploratory use cases, it is important to detect the number of clusters automatically -- a property that most deep clustering methods do not have.
Thus, instead of listing every possible clustering and computing its stability or aiming for a certain number of clusters, we find the most stable clustering similarly as \cite{hdbscan} with a two-step approach where we only need one bottom-up pass through the structure tree followed by a top-down pass: 

\textit{Bottom-up:} 
We regard a cluster $c_a$ implied by a common ancestor $a$. If $a$ is a leaf node of $\mathcal{T}$, we flag the node $a$ as a base case containing at least $2 \mu -2$ points. 
Else, if the stability $S(c_a)$ of $c_a$ is larger than the sum of its children's stabilities, we also flag the node $a$: the cluster $c_a$ leads to higher overall stability than choosing the clusters implied by $a$'s children and is thus preferable: 
If $S(c_a) > \sum_{b \in ch(a)} S(c_{b}) $ then $c_a$ is flagged. 
If not, we store the best possible stability so far in $c_a$ and continue, i.e., $S(c_a):= \sum_{b \in ch(a)} S(c_{b})$.

\textit{Top-down:} Every flagged node with no flagged ancestor defines a cluster. For this, we can simply go through the tree in a depth-first pass and return whenever we hit a flagged node. Nodes that are not flagged and have no flagged ancestor are noise.

\section{Experiments}\label{sec:experiments}
We describe the setup, data, and evaluation in \Cref{sec:setup}, perform a systematic analysis on synthetic data in \Cref{sec:exp-synth}, perform an ablation study in \Cref{sec:exp-ablation}, show our method's quality on benchmark data in \Cref{sec:performance_benchmark}, and discuss its limitations in \Cref{sec:limitation}.

\subsection{Experiment Details}\label{sec:setup}

\paragraph{Algorithms}
We compare SHADE with the following deep clustering methods (cf. \Cref{sec:relatedWork}): The $k$-means like deep clustering methods DCN \cite{DCN}, DEC \cite{DEC}, and IDEC \cite{IDEC}; DipDECK \cite{dipdeck} and DipEncoder \cite{dipencoder}, deep clustering methods that are more flexible w.r.t. cluster shapes; DDC \cite{DDC_deepDensity} a sequential deep density-based clustering method. Note that except for SHADE, DipDECK, and DDC, all the other algorithms need the ground truth number of clusters given by the user in advance.

\paragraph{Setup}
All methods were trained with a batch size of 500, a target embedding size of 10, Adam as the optimizer, and 50 pretrain epochs for the used AE for all methods, except for our method, which does not need a pretrained AE. All methods were then (further) optimized on their respective loss function for 100 epochs. We report average results over ten runs for each experiment. For preprocessing, we applied a z-normalization for all image data and a feature-wise z-normalization for all tabular data. 
For SHADE, we consistently use $\mu=5$ for every experiment. As for the other methods, we used the default settings of the ClustPy package implementation \cite{clustpy} for all their respective hyperparameters. 
Our full code is available at \url{https://github.com/pasiweber/SHADE}.

\paragraph{Data}
We evaluate all algorithms on synthetic data (e.g., Synth\_low, Synth\_high) generated and accessible by the high\-/dimensional density-connected data generator DENSIRED \cite{densired}, tabular data from UCI \cite{uciRepo} (HAR, letterrecognition, htru2, Mice, TCGA, Pendigits), video data (Weizmann \cite{weizmannData}, Keck \cite{keckData}) and image data from UCI \cite{uciRepo} (Coil20, Coil100, cmu\_faces). A list of all datasets, including their properties, can be found in Appendix \ref{appendix:datasets}. Furthermore, we also applied SHADE on popular (deep) clustering benchmark datasets whose clusters usually are of Gaussian shape, as Optdigits \cite{uciRepo}, USPS \cite{usps}, MNIST \cite{mnist}, FMNIST \cite{fmnist}, and KMNIST \cite{kmnist}.

\paragraph{Evaluation} As SHADE is the first deep density-based clustering method that handles noise, the comparison to methods that do not detect noise is not trivial. Hence, in all tables and figures where we compare the results of our algorithm to other methods, we assign every detected noise point to its closest cluster. We label this comparison variant 'SHADE\_1nn'. 
However, assigning every point to a cluster, instead of allowing noise, is neither our goal nor meaningful in exploratory data analysis. This assignment serves the sole purpose of enabling a fair, non-biased assessment of the clustering quality compared to our competitors according to best practices~\cite{overoptimism}.
We report the ARI and NMI values of non-noise points and our algorithm's detected noise percentage in Appendix \ref{appendix:extended_table}.

\begin{table*}[!b]
\centering
\renewcommand\thetable{II}
\caption{SHADE with different batch sizes. The default value is $batch size = 500$ (gray column).
}
\label{tab:ablationBatchsize}
\resizebox{0.84\linewidth}{!}{
\begin{tabular}{l|l|ccccccccc}
\toprule
\textbf{Dataset} & \textbf{Metric} & 100 & 200 & 300 & 400 & \cellcolor{gray!20} 500 & 600 & 700 & 800 & 900\\
\midrule
\parbox[t]{2mm}{\multirow{4}{*}{\rotatebox[origin=c]{55}{\small{Synth\_high}}}}
& ARI & $96.8 \pm 1.3$ & $98.4 \pm 1.2$ & $98.5 \pm 0.6$ & $98.3 \pm 1.0$ & \cellcolor{gray!20}$98.4 \pm 0.6$ & $98.1 \pm 1.3$ & $97.9 \pm 1.1$ & $98.3 \pm 1.1$ & $97.7 \pm 1.2$\\
& NMI & $95.3 \pm 2.1$ & $98.3 \pm 1.2$ & $98.0 \pm 0.7$ & $97.9 \pm 1.0$ & \cellcolor{gray!20}$97.9 \pm 0.6$ & $97.8 \pm 1.3$ & $97.4 \pm 1.3$ & $97.8 \pm 1.4$ & $97.1 \pm 1.4$\\
& noise & $4.3 \pm 2.0$ & $2.6 \pm 0.9$ & $2.8 \pm 1.4$ & $3.5 \pm 1.1$ & \cellcolor{gray!20}$3.0 \pm 1.7$ & $3.0 \pm 1.2$ & $4.1 \pm 1.4$ & $3.8 \pm 1.7$ & $3.1 \pm 1.1$\\
& $k$ & $20.6 \pm 5.1$ & $12.6 \pm 2.2$ & $13.7 \pm 1.4$ & $13.4 \pm 1.6$ & \cellcolor{gray!20}$13.7 \pm 1.5$ & $13.1 \pm 1.6$ & $14.1 \pm 2.5$ & $13.7 \pm 2.2$ & $14.4 \pm 1.8$\\
\midrule
\parbox[t]{2mm}{\multirow{4}{*}{\rotatebox[origin=c]{55}{\small{TCGA}}}}
& ARI & $86.7 \pm 7.2$ & $84.5 \pm 4.9$ & $82.8 \pm 6.0$ & $83.5 \pm 7.9$ & \cellcolor{gray!20}$85.8 \pm 6.2$ & $84.8 \pm 9.7$ & $88.2 \pm 4.6$ & $90.2 \pm 7.0$ & $90.0 \pm 4.2$\\
& NMI & $91.3 \pm 4.3$ & $89.3 \pm 3.4$ & $87.6 \pm 4.6$ & $88.7 \pm 5.3$ & \cellcolor{gray!20}$89.5 \pm 3.8$ & $89.0 \pm 6.3$ & $91.6 \pm 3.2$ & $93.5 \pm 4.0$ & $93.5 \pm 2.5$\\
& noise & $22.3 \pm 5.0$ & $28.2 \pm 4.5$ & $23.3 \pm 9.2$ & $27.6 \pm 6.6$ & \cellcolor{gray!20}$27.3 \pm 5.7$ & $25.8 \pm 9.4$ & $26.2 \pm 7.9$ & $24.5 \pm 7.0$ & $22.1 \pm 5.1$\\
& $k$ & $6.4 \pm 0.9$ & $6.1 \pm 1.4$ & $5.7 \pm 1.0$ & $6.2 \pm 1.4$ & \cellcolor{gray!20}$6.0 \pm 1.0$ & $6.1 \pm 0.9$ & $5.8 \pm 0.6$ & $5.2 \pm 1.0$ & $6.3 \pm 0.8$\\
\midrule
\parbox[t]{2mm}{\multirow{4}{*}{\rotatebox[origin=c]{55}{\small{COIL20}}}}
& ARI & $77.6 \pm 8.4$ & $79.6 \pm 7.8$ & $84.1 \pm 5.0$ & $84.3 \pm 3.3$ & \cellcolor{gray!20}$85.2 \pm 1.9$ & $82.4 \pm 4.2$ & $82.7 \pm 4.3$ & $84.2 \pm 3.2$ & $83.2 \pm 4.8$\\
& NMI & $91.3 \pm 2.6$ & $92.6 \pm 2.1$ & $94.0 \pm 1.6$ & $94.3 \pm 1.1$ & \cellcolor{gray!20}$94.3 \pm 1.0$ & $93.6 \pm 1.2$ & $93.7 \pm 1.4$ & $94.0 \pm 1.4$ & $93.6 \pm 1.7$\\
& noise & $14.0 \pm 3.4$ & $15.0 \pm 3.9$ & $15.7 \pm 2.1$ & $13.1 \pm 2.6$ & \cellcolor{gray!20}$13.6 \pm 2.0$ & $13.2 \pm 2.5$ & $14.8 \pm 1.9$ & $13.9 \pm 1.3$ & $13.0 \pm 2.2$\\
& $k$ & $17.9 \pm 2.1$ & $17.4 \pm 1.3$ & $17.9 \pm 1.6$ & $17.1 \pm 1.2$ & \cellcolor{gray!20}$16.9 \pm 0.7$ & $16.5 \pm 1.2$ & $16.7 \pm 1.6$ & $16.7 \pm 0.6$ & $16.7 \pm 0.9$\\
\bottomrule
\end{tabular}}
\end{table*}

\begin{figure}[!b]
    \centering
    \includegraphics[width=.99\linewidth]{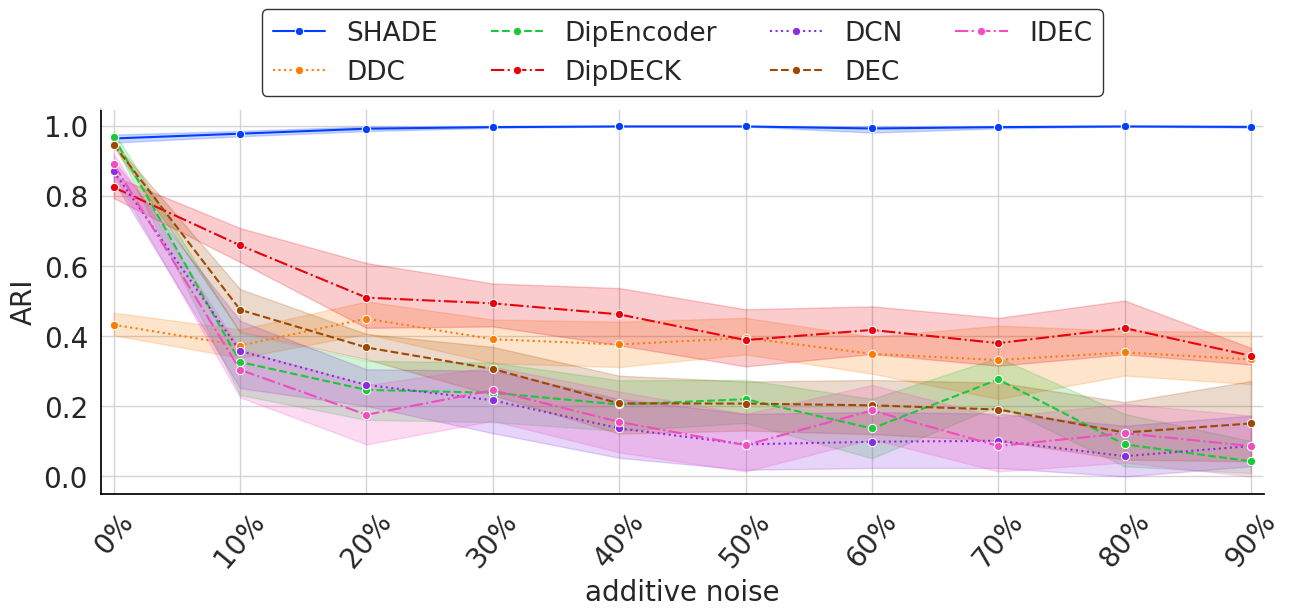}%
    \caption{ARI on synthetic data with varying noise ratio and 100 dimensions. SHADE consistently surpasses all competitors. Note that SHADE reliably returns the highest ARI values, whereas our competitors have a high variance in clustering quality across ten runs. This shows the importance of inherent noise handling for deep clustering algorithms.
    }\label{fig:exp-noise}
\end{figure}

\subsection{Comparison of Competitors on Toy Example} To provide some intuition about the strengths and weaknesses of current deep clustering methods, let us regard our motivational example in \Cref{fig:motivation}:
The attempt of DEC to force all clusters to Gaussian blobs in 2D space fails as Figure \ref{fig:motivation}(d) demonstrates: The two rings are not well separated, and the clusters are transformed into Gaussian blobs, losing all of its original structure.
Combining the centroid-based clustering loss with the reconstruction loss of the AE, IDEC preserves most of the structure of the s-shaped cluster but fails to separate the rings (cf. \ref{fig:motivation}(e)). 
DCN \cite{DCN} shows a very similar behavior. Although it fails to separate the two rings, it preserves the overall structure quite well, as seen in Figure 1(f). As seen in Figure 1(g), the DipEncoder \cite{dipencoder} completely loses the original structure of the clusters as it tries to achieve an unimodal distribution within the embedded clusters. However, this strategy is incompatible with circular or s-shaped clusters, as these always have at least two modes regardless of the perspective. DipDECK \cite{dipdeck} can also not separate the rings and also splits up one ring, as seen in Figure 1(h).
DDC \cite{DDC_deepDensity} clusters on the TSNE-embedding of an already trained AE embedding. Figure 1(i) shows that even this two-stage process fails to separate the two rings. However, it mostly preserves the density structure of the clusters.

\subsection{Systematic Evaluation regarding Noise}
\label{sec:exp-synth}
We systematically evaluate our algorithm SHADE against the competitor methods w.r.t. varying levels of \textit{noise}. We generated multiple synthetic datasets with $5,000$ data points and $100$ dimensions with the data generator DENSIRED \cite{densired}. We let the data generator add additive, uniform noise up to 90\% and investigated the clustering performance on the non-noise points. As shown in \Cref{fig:exp-noise}, SHADE is the only deep clustering algorithm that can still cluster the non-noise points correctly. All the other methods are affected by the additional noise points, and their cluster performance drops drastically.

\subsection{Ablation Study}\label{sec:exp-ablation}

\subsubsection{Varying Parameter $\mu$}

\begin{table}
\centering
\renewcommand\thetable{I}
\caption{SHADE's clustering results for varying $\mu$. \\
The default value for SHADE is $\mu = 5$ (gray column).
}\label{tab:minptsAblation}
\resizebox{0.99\linewidth}{!}{
\begin{tabular}{l|l|ccccc}
\toprule
\textbf{Dataset} & \textbf{Metric} & $\mu = 3$ & $\mu = 4$ & \cellcolor{gray!20} $\mu = 5$ & $\mu = 6$ & $\mu = 7$\\
\midrule
\parbox[t]{2mm}{\multirow{4}{*}{\rotatebox[origin=c]{55}{\small{Synth\_high}}}}
& ARI & $97.5 \pm 1.1$ & $98.6 \pm 1.1$ & \cellcolor{gray!20} $98.4 \pm 1.2$ & $98.3 \pm 0.9$ & $98.0 \pm 1.5$\\
& NMI & $97.0 \pm 1.1$ & $98.2 \pm 1.2$ & \cellcolor{gray!20} $98.1 \pm 1.2$ & $98.1 \pm 0.7$ & $97.4 \pm 1.7$\\
& noise & $3.1 \pm 1.4$ & $2.8 \pm 0.9$ & \cellcolor{gray!20} $2.4 \pm 1.2$ & $3.1 \pm 0.9$ & $3.2 \pm 0.9$\\
& $k$ & $14.6 \pm 1.7$ & $13.0 \pm 1.8$ & \cellcolor{gray!20} $13.0 \pm 1.5$ & $12.9 \pm 0.7$ & $14.4 \pm 2.7$\\
\midrule
\parbox[t]{2mm}{\multirow{4}{*}{\rotatebox[origin=c]{55}{\small{TCGA}}}}
& ARI & $79.5 \pm 11.5$ & $84.1 \pm 8.2$ & \cellcolor{gray!20} $86.6 \pm 7.8$ & $85.2 \pm 6.7$ & $84.7 \pm 6.6$\\
& NMI & $86.8 \pm 6.2$ & $89.2 \pm 5.5$ & \cellcolor{gray!20} $90.8 \pm 4.4$ & $89.3 \pm 5.2$ & $89.6 \pm 4.0$\\
& noise & $20.6 \pm 8.8$ & $22.1 \pm 7.7$ & \cellcolor{gray!20} $21.8 \pm 9.0$ & $21.6 \pm 2.8$ & $23.5 \pm 8.4$\\
& $k$ & $5.8 \pm 1.7$ & $6.2 \pm 1.2$ & \cellcolor{gray!20} $5.7 \pm 0.9$ & $5.8 \pm 1.0$ & $6.1 \pm 0.7$\\
\midrule
\parbox[t]{2mm}{\multirow{4}{*}{\rotatebox[origin=c]{55}{\small{COIL20}}}}
& ARI & $83.6 \pm 4.8$ & $81.3 \pm 8.4$ & \cellcolor{gray!20} $82.5 \pm 4.5$ & $84.5 \pm 3.1$ & $83.4 \pm 4.2$\\
& NMI & $94.2 \pm 1.7$ & $93.3 \pm 2.0$ & \cellcolor{gray!20} $93.6 \pm 1.7$ & $94.2 \pm 1.1$ & $93.9 \pm 1.0$\\
& noise & $11.4 \pm 2.2$ & $10.9 \pm 2.2$ & \cellcolor{gray!20} $12.5 \pm 1.9$ & $14.2 \pm 1.8$ & $14.8 \pm 3.4$\\
& $k$ & $16.8 \pm 0.6$ & $16.6 \pm 0.8$ & \cellcolor{gray!20} $16.5 \pm 1.0$ & $16.9 \pm 0.8$ & $17.3 \pm 0.8$\\
\bottomrule
\end{tabular}}
\end{table}

Density-connectivity often depends on hyperparameters like $\mu$ and $\varepsilon$. 
As $\mathcal{T}_d$ captures all possible density-connected clusterings in dependence of $\varepsilon$, users do not need to predefine $\varepsilon$. 
For choosing $\mu$, there are several heuristics, e.g., $\mu > d$ is often advised, and one of the most recent and widespread heuristics~\cite{dbscan-rev-rev} is $\mu = 2d-1$.
Note that this is aimed at datasets with significantly more points than dimensions, which can no longer be considered the default. 
The parameter $\mu$ is mainly needed for robustness and countering the single link effect that merges different clusters because of chains of noise in between them~\cite{chaindetection}. 
However, already in the original DBSCAN paper~\cite{dbscan}, it is set to $\mu = 4$ for all experiments, as it does not significantly change the results. 
This aligns with established~\cite{wishart_density_SL} and recent research~\cite{dbscan-rev-rev}.

We investigated SHADE's sensitivity to the parameter $\mu$ and show that the clustering quality, noise detection, and the estimation of the number of clusters $k$ do not vary much for different values of $\mu$. We present the results of our algorithm for $\mu \in [3,4,5,6,7]$, demonstrated on three of the benchmark datasets in \Cref{tab:minptsAblation}. 
We note that for very small datasets, like TCGA with only 801 points, the choice of $\mu$ can cause slightly varying clustering results. However, for the typical deep learning setting, clustering quality is robust w.r.t. $\mu$. 
Thus, we set a low value for $\mu = 5$ for \textit{all experiments}, as this is enough to avoid single links, and similar values are often recommended in the literature~\cite{dbscan-rev-rev, dbscan}. 

\begin{table}[!tb]

\centering
\renewcommand\thetable{III}
\caption{Individual performance and the average score of SHADE on some datasets with varying loss functions.\\
Per default, both $\mathcal{L}_{rec}$ and $\mathcal{L}_d$ are included (gray column).
}
\label{tab:ablation_loss}
\resizebox{0.77\linewidth}{!}{
\begin{tabular}{l|l|c|c|c}
\toprule
\textbf{Dataset} & \textbf{Metric} & \cellcolor{gray!20} $\mathcal{L}_D + \mathcal{L}_{rec}$ & $\mathcal{L}_D$ & $\mathcal{L}_{rec}$ \\
\midrule
\parbox[t]{2mm}{\multirow{3}{*}{\rotatebox[origin=c]{40}{\footnotesize{Synth\_low}}}}  
& ARI & \cellcolor{gray!20} \bm{$99.7 \pm 0.5$} & $86.6 \pm 0.4$ & \underline{$97.1 \pm 3.3$}\\
& NMI & \cellcolor{gray!20} \bm{$99.6 \pm 0.4$} & $90.4 \pm 0.1$ & \underline{$98.2 \pm 1.5$}\\
& DCSI & \cellcolor{gray!20} \underline{$97.5 \pm 0.5$} & \bm{$98.8 \pm 0.2$} & $97.1 \pm 0.5$\\
\midrule 
\parbox[t]{2mm}{\multirow{3}{*}{\rotatebox[origin=c]{40}{\footnotesize{HAR}}}}  
& ARI & \cellcolor{gray!20} \bm{$37.8 \pm 7.2$} & $33.0 \pm 4.8$ & \underline{$33.2 \pm 0.0$}\\
& NMI & \cellcolor{gray!20} \bm{$58.8 \pm 5.7$} & $52.1 \pm 2.8$ & \underline{$55.6 \pm 0.1$}\\
& DCSI & \cellcolor{gray!20} \bm{$48.3 \pm 5.2$} & $23.7 \pm 2.5$ & \underline{$43.8 \pm 7.9$}\\
\midrule
\parbox[t]{2mm}{\multirow{3}{*}{\rotatebox[origin=c]{40}{\footnotesize{letterrec.}}}}  
& ARI & \cellcolor{gray!20} \underline{$23.0 \pm 2.0$} & \bm{$23.3 \pm 1.2$} & \underline{$23.0 \pm 1.2$}\\
& NMI & \cellcolor{gray!20} \underline{$57.6 \pm 0.7$} & \bm{$58.1 \pm 0.6$} & $56.1 \pm 0.6$\\
& DCSI & \cellcolor{gray!20} \bm{$23.0 \pm 1.2$} & $19.4 \pm 1.7$ & \underline{$22.0 \pm 1.6$}\\
\midrule
\parbox[t]{2mm}{\multirow{3}{*}{\rotatebox[origin=c]{40}{\footnotesize{TCGA}}}}  
& ARI & \cellcolor{gray!20} \underline{$82.3 \pm 9.9$} & $53.7 \pm 13.4$ & \bm{$82.9 \pm 6.0$}\\
& NMI & \cellcolor{gray!20} \bm{$88.8 \pm 5.3$} & $69.5 \pm 11.6$ & \underline{$88.5 \pm 3.1$}\\
& DCSI & \cellcolor{gray!20} \bm{$77.3 \pm 3.7$} & $60.5 \pm 1.3$ & \underline{$66.0 \pm 3.2$}\\
\midrule
\midrule
\parbox[t]{2mm}{\multirow{3}{*}{\rotatebox[origin=c]{40}{\footnotesize{Average}}}}  
& ARI & \cellcolor{gray!20} \bm{$60.7 \pm 4.9$} & \underline{$59.0 \pm 2.6$} & $49.2 \pm 4.9$ \\
& NMI & \cellcolor{gray!20} \bm{$76.2 \pm 3.0$} & \underline{$74.6 \pm 1.3$} & $67.5 \pm 3.8$ \\
& DCSI & \cellcolor{gray!20} \bm{$	61.5 \pm 2.7$} & \underline{$57.2 \pm 3.3$} & $50.6 \pm 1.4$ \\
\bottomrule
\end{tabular}}
\end{table}

\subsubsection{Varying Batch Size}\label{sec:ablation_batch_size}
The batch size in the training phase of our algorithm is more important than in other deep clustering algorithms. By using batches, we only approximately optimize our original loss function stated in \Cref{eq:DC_Loss_orig}. Hence, we conducted an ablation study on different batch sizes. As shown in \Cref{tab:ablationBatchsize}, our algorithm is stable over a wide range of different batch sizes. Sometimes, a larger batch size can improve performance, as seen on the TCGA dataset, but generally, the performance is similar for any batch size that is not too small (e.g., $\geq 200$ for Synth\_high or $\geq 300$ for COIL20). Hence, we choose a batch size of $500$ in the comparison experiment with our competitors. 

\renewcommand{\arraystretch}{1.1}

\begin{table*}[ht]

\centering
\caption{Clustering results measured by ARI of SHADE\_1nn, where we assign all noise points to their nearest cluster for better comparability with our competitors. Note that we discuss results for TCGA in \Cref{sec:performance_benchmark} and explain the performance for HAR in more detail in \Cref{sec:limitation} and \Cref{fig:harConfMatrix}.}
\label{tab:experiments_ari}
\resizebox{0.85\linewidth}{!}{
\begin{tabular}{rl|ccccccc|c}
\toprule
& \textbf{Dataset} & SHADE\_1nn & DDC & DipDECK & DipEncoder & DCN & DEC & IDEC & src \\
\midrule
\multirow{8}{*}{\rotatebox[origin=c]{90}{Tabular data}}
& Synth\_low & \cellcolor{Green!70}\bm{$98.9 \pm 2.0$} & \cellcolor{Green!39}\underline{$56.9 \pm 5.5$} & \cellcolor{Green!23}$33.9 \pm 6.4$ & \cellcolor{Green!5}$10.1 \pm 9.9$ & \cellcolor{Green!5}$9.6 \pm 9.7$ & \cellcolor{Green!27}$40.2 \pm 3.0$ & \cellcolor{Green!9}$15.3 \pm 8.8$ & 
\multirow{2}{*}{\rotatebox[origin=c]{0}{\cite{densired}}}
\\
& Synth\_high & \cellcolor{Green!70}\bm{$97.5 \pm 1.4$} & \cellcolor{Green!23}\underline{$33.9 \pm 11.1$} & \cellcolor{Green!20}$29.9 \pm 13.6$ & \cellcolor{Green!5}$9.3 \pm 10.7$ & \cellcolor{Green!5}$8.8 \pm 10.5$ & \cellcolor{Green!21}$30.3 \pm 3.5$ & \cellcolor{Green!12}$17.9 \pm 6.6$
\\ \cmidrule{2-10}
& HAR & \cellcolor{Green!5}$36.4 \pm 6.4$ & \cellcolor{Green!33}$49.4 \pm 3.3$ & \cellcolor{Green!38}$51.3 \pm 4.0$ & \cellcolor{Green!57}$60.0 \pm 6.9$ & \cellcolor{Green!70}\bm{$66.1 \pm 1.3$} & \cellcolor{Green!64}$63.4 \pm 2.4$ & \cellcolor{Green!67}\underline{$64.9 \pm 0.9$} &
\multirow{6}{*}{\rotatebox[origin=c]{0}{\cite{uciRepo}}}
\\
& letterrec. & \cellcolor{Green!62}$23.0 \pm 0.9$ & \cellcolor{Green!14}$9.9 \pm 2.9$ & \cellcolor{Green!5}$7.3 \pm 3.5$ & \cellcolor{Green!68}\underline{$24.7 \pm 1.3$} & \cellcolor{Green!61}$22.6 \pm 1.0$ & \cellcolor{Green!65}$23.9 \pm 1.6$ & \cellcolor{Green!70}\bm{$25.2 \pm 1.8$}\\
& htru2 & \cellcolor{Green!70}\bm{$65.0 \pm 19.5$} & \cellcolor{Green!54}$49.4 \pm 13.0$ & \cellcolor{Green!12}$9.7 \pm 19.4$ & \cellcolor{Green!6}$4.3 \pm 0.8$ & \cellcolor{Green!54}\underline{$49.7 \pm 2.8$} & \cellcolor{Green!5}$3.0 \pm 0.5$ & \cellcolor{Green!5}$3.2 \pm 0.6$\\
& Mice & \cellcolor{Green!70}\bm{$27.7 \pm 2.9$} & \cellcolor{Green!43}\underline{$25.2 \pm 1.9$} & \cellcolor{Green!17}$22.7 \pm 4.3$ & \cellcolor{Green!5}$21.6 \pm 2.6$ & \cellcolor{Green!6}$21.7 \pm 1.4$ & \cellcolor{Green!9}$22.0 \pm 1.5$ & \cellcolor{Green!7}$21.8 \pm 1.4$\\
& TCGA & \cellcolor{Green!5}$80.0 \pm 13.7$ & \cellcolor{Green!41}$87.5 \pm 0.8$ & \cellcolor{Green!48}\underline{$88.8 \pm 4.4$} & \cellcolor{Green!70}\bm{$93.4 \pm 6.0$} & \cellcolor{Green!40}$87.2 \pm 5.3$ & \cellcolor{Green!30}$85.1 \pm 2.7$ & \cellcolor{Green!18}$82.6 \pm 0.9$\\
& Pendigits & \cellcolor{Green!62}\underline{$75.1 \pm 0.8$} & \cellcolor{Green!70}\bm{$76.9 \pm 2.0$} & \cellcolor{Green!59}$74.3 \pm 1.1$ & \cellcolor{Green!18}$64.6 \pm 3.0$ & \cellcolor{Green!5}$61.6 \pm 1.9$ & \cellcolor{Green!22}$65.7 \pm 3.3$ & \cellcolor{Green!19}$64.9 \pm 2.6$\\
\midrule
\multirow{2}{*}{\rotatebox[origin=c]{90}{Video}}
& Weizmann & \cellcolor{Green!70}\bm{$48.2 \pm 3.6$} & \cellcolor{Green!10}$14.7 \pm 1.8$ & \cellcolor{Green!5}$12.0 \pm 1.9$ & \cellcolor{Green!25}$23.3 \pm 1.2$ & \cellcolor{Green!28}$24.6 \pm 1.1$ & \cellcolor{Green!28}\underline{$24.9 \pm 1.2$} & \cellcolor{Green!28}$24.7 \pm 1.2$& \cite{weizmannData} \\
& Keck & \cellcolor{Green!70}\bm{$7.5 \pm 0.4$} & \cellcolor{Green!5}$-0.2 \pm 1.1$ & \cellcolor{Green!65}$6.9 \pm 0.8$ & \cellcolor{Green!67}\underline{$7.1 \pm 0.3$} & \cellcolor{Green!61}$6.4 \pm 0.5$ & \cellcolor{Green!58}$6.1 \pm 0.9$ & \cellcolor{Green!59}$6.2 \pm 0.9$ & \cite{keckData}\\
\midrule
\multirow{3}{*}{\rotatebox[origin=c]{90}{Image}}
& COIL20 & \cellcolor{Green!70}\bm{$68.7 \pm 3.5$} & \cellcolor{Green!46}$62.0 \pm 5.5$ & \cellcolor{Green!5}$50.5 \pm 7.8$ & \cellcolor{Green!53}\underline{$64.0 \pm 3.0$} & \cellcolor{Green!47}$62.4 \pm 2.8$ & \cellcolor{Green!52}$63.7 \pm 2.8$ & \cellcolor{Green!49}$62.9 \pm 2.9$ &
\multirow{3}{*}{\rotatebox[origin=c]{0}{\cite{uciRepo}}}
\\
& COIL100 & \cellcolor{Green!70}\underline{$56.8 \pm 5.0$} & \cellcolor{Green!5}$16.4 \pm 3.8$ & \cellcolor{Green!13}$21.4 \pm 3.0$ & \cellcolor{Green!66}$54.3 \pm 1.9$ & \cellcolor{Green!68}$55.9 \pm 3.0$ & \cellcolor{Green!68}$55.8 \pm 2.0$ & \cellcolor{Green!70}\bm{$56.9 \pm 2.0$} & \\
& cmu\_faces & \cellcolor{Green!35}$34.6 \pm 6.2$ & \cellcolor{Green!37}$35.0 \pm 3.5$ & \cellcolor{Green!5}$29.8 \pm 9.8$ & \cellcolor{Green!55}$37.9 \pm 2.2$ & \cellcolor{Green!70}\bm{$40.3 \pm 2.0$} & \cellcolor{Green!42}$35.8 \pm 2.8$ & \cellcolor{Green!64}\underline{$39.4 \pm 3.3$} & \\
\bottomrule
\end{tabular}}
\end{table*}

\subsubsection{Different Parts of our Loss Function}
We performed an ablation study regarding the effect of the different parts of our loss function. For this, we apply the clustering step explained in \Cref{sec:fullyAutomaticClustering} in three different embeddings resulting from different loss functions: the loss $\mathcal{L}_{D}$ combined with $\mathcal{L}_{rec}$, only the $\mathcal{L}_{D}$ loss, and only the $\mathcal{L}_{rec}$ loss. \Cref{tab:ablation_loss} shows the results on a selection of datasets and the average results of them. The individual results on each dataset are provided in Appendix \ref{appendix:loss_extended}. On average, the combination of the $\mathcal{L}_{D}$ and $\mathcal{L}_{rec}$ loss performs best, followed by using only our novel $\mathcal{L}_{D}$ loss. 
This holds true for all three evaluation measures: ARI and NMI indicate the correspondence to ground truth clusters. DCSI captures the separation between ground truth clusters and the connectedness within ground truth clusters in the respective embedding. A high DCSI indicates that the ground truth clusters can be found in the embedding using a density-based clustering approach.

\subsection{Performance Benchmark}\label{sec:performance_benchmark}
Our benchmark study results are shown in \Cref{tab:experiments_ari}, where we used the competitors and datasets described in \Cref{sec:setup}. For most datasets, SHADE has the best or second-best quality. 

One exception is the TCGA dataset, which contains five different tumor types as clusters. Based on the good results of centroid-based methods in \Cref{tab:experiments_ari}, we believe that TCGA contains Gaussian-like clusters, which are well-represented with prototypes for each tumor type.
On that note, we want to emphasize that we tested data that \emph{potentially} contains density-connected structures -- frequently used benchmark datasets like MNIST do not contain \textit{density-connected} clusters, but spherical-shaped ones and are thus not listed here. To find such Gaussian clusters, we recommend using existing centroid-based deep clustering algorithms. Nevertheless, we provide the results of our algorithm on various MNIST versions in Appendix \ref{appendix:mnist}.

\Cref{tab:experiments_ari} shows that SHADE is remarkably successful on the datasets where we know that they contain density-based structures, i.e., the synthetic datasets and the Weizmann video data. On synthetic datasets, our algorithm reaches an ARI above $97.0$, whereas the second-best method merely reaches an ARI of $56.9$ resp. $33.9$. Not surprisingly, for both datasets, the runner-up is DDC -- our only competitor including a concept of density. 
We provide further information like NMI values and the detected number of clusters in Appendix \ref{appendix:extended_table}.

\subsection{Limitations}\label{sec:limitation}
\begin{figure}[!htb]
    \centering
    \includegraphics[width=0.87\linewidth]{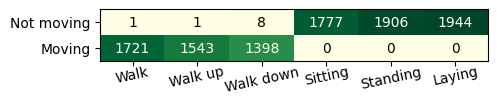}
    \caption{HAR Confusion Matrix. x-axis gives ground truth classes, y-axis clusters  detected by SHADE}
    \label{fig:harConfMatrix}
\end{figure}

We showed superior performance on data containing density\-/connected structures in the high-dimensional space. 
However, like any clustering algorithm, SHADE is not perfect in all cases. An illustrative example is the Human Action Recognition (HAR) dataset. 
Our competitors performed consistently better than SHADE, especially for the methods where the number of clusters is given. 
However, the confusion matrix in \Cref{fig:harConfMatrix} and a brief analysis of the class labels reveal the reason behind this behavior: the dataset contains motion data for six different activities. However, three of them are activities where people move, and the others are resting activities. These classes are density-connected, as the transition between the activities \emph{walk}, \emph{walk upstairs}, and \emph{walk downstairs} are smooth. SHADE, thus, almost perfectly differentiates between moving and not moving activities, which might not yield competitive ARI values but is a reasonable clustering nevertheless. 

\section{Discussion and Related Work}\label{sec:relatedWork}
In the following, we provide some background on methods that detect density-based clusters and on current existing deep clustering methods. Furthermore, we discuss the existing approach that applies density-based clustering on the resulting representation of applying several dimensional reduction techniques sequentially.

\subsection{Density-based Clustering}\label{sec:relwork-density}%
We distinguish two different types of density-based clustering methods: 1) Those that define a cluster as one dense area that has a certain density, i.e., points per volume (e.g., Density Peaks~\cite{density-peaks}).
2) Those that rely on the concept of \textit{density-connectivity} as known from DBSCAN-like \cite{dbscan} methods. 
This well-established concept, first introduced in the 70s~\cite {wishart_density_SL, hartigan1975clustering}, serves as the basis for SHADE.

Density-based algorithms have several important advantages compared to other clustering paradigms. Most of them can find clusters of arbitrary shape, detect the number of clusters automatically, and can handle noise. These are invaluable properties for real-world use cases, e.g., for pioneering research in science. In these cases, the expected results or properties of the data are usually not known beforehand. 
Where centroid-based clustering methods are often based on strong assumptions about the data, like assuming underlying Gaussian distributions (e.g., the EM-algorithm~\cite{em}), density-based methods are more flexible and data-driven. The associated hyperparameters, however, are often hard to set.

Furthermore, density-based clustering methods are often inherently slow as they are based on complex graph-based computations (even if they are only implicitly used).
The \textit{curse of dimensionality}~\cite{curseofdim} is another issue that makes finding density-based structures in high-dimensional space hard, as the
pairwise distances between points become increasingly similar. Thus, the inter- and intra-cluster distances also become harder to distinguish.
The \textit{empty space problem}~\cite{steinbach2004challenges} consolidates with increasing dimensionality, making dense areas in the data space more and more unlikely. 
Thus, finding density-connected structures is, in general, no trivial task, and even though density-based clustering is an established concept, it is still highly discussed, and the research community constantly advances its knowledge \cite{dbscan-rev-rev, dcdist}.

\subsection{Deep Clustering}
Deep clustering describes the combination of deep learning techniques with specific clustering objectives. Here, various deep learning architectures can be used, with most deep clustering methods based on feedforward \cite{feedforwardAE} or convolutional \cite{convolutionalAE} autoencoders (AEs). AEs transform the data into a lower-dimensional space by applying the encoding function $enc(\cdot)$ and try to reconstruct the original representation using the decoding function $dec(\cdot)$. Deep clustering approaches utilizing AEs usually optimize the loss function $\mathcal{L} = \lambda_1 \mathcal{L}_{rec} + \lambda_2 \mathcal{L}_{clust}$, where $\mathcal{L}_{rec}$ validates the quality of the embedding by comparing the input $x$ and output $enc(dec(x))$ of the AE and $\mathcal{L}_{clust}$ strengthens the existing clustering structures.
Most established deep clustering methods use a centroid-based objective as the basis for  $\mathcal{L}_{clust}$. They are often related to $k$-Means using either soft assignments, like DEC \cite{DEC} and IDEC \cite{IDEC} or hard assignments, like DCN \cite{DCN}, DKM \cite{DKM} and ACe/DeC \cite{Acedec} for centroid-based deep clustering. The centroid-based objective pulls clusters together such that they can collapse onto their cluster centroid. While this can be beneficial for clustering accuracy, it completely destroys the density-based structure of a dataset. Another problem with these approaches is that they assume that distances within the clusters are smaller than distances between clusters - an assumption that is not true for density-connected clusters. 
Other methods attempt to address this problem.
For example, the DipEncoder \cite{dipencoder} considers modalities instead of distances. Therefore, each cluster can obtain an individual spread, leading to more diverse patterns within the embedding.
Despite the increased flexibility, it can still only identify convex cluster shapes.
DipDECK \cite{dipdeck} overcomes this limitation by initially overestimating the number of clusters, which are later merged if they show a coherent structure. Here, multiple $ k$-means-like clusters can be combined to capture more complex patterns. Furthermore, this strategy allows them to estimate the number of clusters.

\subsection{Sequential Deep Density-based Clustering} 
Current methods at the intersection of deep and density-based clustering only work sequentially. They first learn a representation (independently of the density-based clustering objective) using a neural network and, afterward, apply a (non-deep) density-based clustering method on top of this representation. This stands in strong contrast to our method and the deep clustering methods described above, which optimize the clustering and representation simultaneously with a unified objective. 
An example of a sequential deep density-based clustering method is DDC (Deep Density-based image Clustering) \cite{DDC_deepDensity}. It performs three steps. First, an autoencoder is pretrained to obtain a lower-dimensional representation. Second, t-SNE \cite{tsne} is employed to further reduce the dimensionality. Third, a variant of Density-Peak Clustering (DPC) \cite{density-peaks} is employed to obtain cluster assignments from the doubly reduced representation. This process allows DDC to exploit the advantages of both transformations, but at the same time, it inherits the disadvantages. If the AE learns a poor initial embedding, undesired structures are reinforced by t-SNE, resulting in an insufficient final representation. In addition, outliers can strongly distort the result of t-SNE. As DDC follows a sequential, three-step approach, it neglects the integration of density-based clustering and deep learning under a common objective and, therefore, can not improve from a bad representation.
Note that by applying DPC as a last step, found clusters are density-\textit{based}, but this does not imply that the algorithm finds density-\textit{connected} structures (see \Cref{sec:relwork-density}): e.g., DDC does not find elongated density-connected structures like in our motivational example in \Cref{fig:motivation}. 
Nevertheless, DDC achieves an overall strong cluster performance, even when used without augmentations. Thus, we use it as a comparison method in our experiments in Section \ref{sec:experiments}.

\subsection{Notions of Stability}\label{sec:rel:stability}
Of the various definitions for the 'stability' of a cluster, most are complex, arbitrary, or come with other severe disadvantages. 
They often rely on sampling or generating perturbed versions of the dataset, which is then clustered and compared to other clustering results~\cite{stability_luxburg}. 
These approaches are computationally expensive, as several clusterings have to be computed, and are usually not deterministic.
E.g., \cite{approximationStability} defines a clustering as stable if all clusterings that yield a similarly good solution w.r.t. the optimization function are similar to it. 
\cite{smith1980stability} rely on random splits of the data and compare clusterings on the parts with those on the full dataset. This is computationally complex as several partitionings need to be computed and compared.
HDBSCAN \cite{hdbscan_long} defines stability based on an Euclidean MST defining the 'death' and 'birth' of clusters. While our definitions have a similar intuition, note that our MST is based on the mutual reachability distance, as it is a closer fit to the original concepts of density-connectivity as given by DBSCAN$^{(*)}$~\cite{dbscan,dcdist}.

\section{Conclusion}
We present SHADE, the first deep clustering method that incorporates density in its loss function. 
It preserves density-connected structures in low-dimensional embeddings, allowing a meaningful visualization that is especially useful for exploratory data analysis.
The embedding furthermore enables SHADE to find arbitrarily shaped clusters and even topologically intertwined structures while detecting the number of clusters fully automatically. We performed extensive experiments on various datasets, showing SHADE's superiority on various datasets and noisy data.

\section*{Acknowledgements}
We gratefully acknowledge financial support from the Vienna Science and Technology Fund (WWTF-ICT19-041) and the Austrian funding agency for business-oriented research, development, and innovation (FFG-903641\footnote{\url{https://projekte.ffg.at/projekt/4814676}}).

\bibliographystyle{IEEEtranS}
\bibliography{lib}


\appendices
\lhead{SHADE: Deep Density-based Clustering -- Appendix}

\section{Density-Connectivity Tree and Structure Tree}\label{appendix:dctree}
\begin{figure}[!hb]
    \centering
    \caption{Tree capturing density-connectivity. Leaves correspond to data points and nodes to the distances $d_{dc}$ between them.  }
    \label{fig:dctree}
\begin{forest}
  for tree={
    circle, 
    draw,
    minimum size=0em,
    edge={->},
    align=center,
    l sep+=0mm,
    s sep+=0mm,
    scale=0.8,
  },
  where n children=0{
    rectangle, 
    draw,
    fill=green!20,
  }{},
  [$n_r$, draw
    [$n_0$, draw
      [$l_1$]
      [$l_2$]
    ]
    [$n_1$, draw,  line width=3pt, fill=red!20 
      [$n_2$, draw,  line width=3pt, fill=blue!20 
        [$n_4$, draw, fill=blue!20
            [$n_7$, draw, fill=blue!20
              [$l_3$]
              [$l_4$]
            ]
            [$n_8$, draw, fill=blue!20
              [$l_5$]
              [$l_6$]
            ]
        ]
        [$n_5$, draw, fill=blue!20
            [$n_{9}$, draw, fill=blue!20
              [$l_7$]
              [$l_8$]
            ]
          [$l_9$]
        ]
      ]
      [$n_{3}$, draw, fill=red!20
        [$l_{10}$]
        [$n_6$, draw, fill=red!20
          [$l_{11}$]
          [$l_{12}$]
        ]   
      ]
    ]
  ]
\end{forest}
\end{figure}
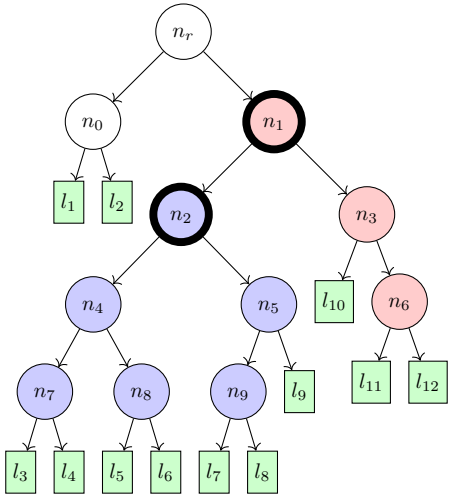

\Cref{fig:dctree} shows a small example of a density-connectivity tree $\mathcal{T}_d$ for $\mu = 3$. 
The nodes with bold borders are nodes that exist in the structure tree, too, as both their children have more than $\mu=3$ leaves. Colors imply the smallest cluster to which a node belongs in the structure tree. 

\newpage

\section{Overview of dataset Properties for Experiments}
\label{appendix:datasets}
\Cref{tab:exps-stats} summarizes all datasets that we used, their number of points $n$, dimensionality $d$, number of classes $k$, number of noise points \#noise, and their source. The noise in Synth\_low and Synth\_high was sampled from a uniform distribution, where noise has to show a certain distance to the clusters.

\begin{table}[!htb]
\caption{Dataset properties}\label{tab:exps-stats}
\resizebox{0.99\linewidth}{!}{
\begin{tabular}{clrrrcr}
\toprule
& Dataset & $n$ & $d$ & $k$ & \#noise & Source \\
\midrule
\parbox[t]{2mm}{\multirow{7}{*}{\rotatebox[origin=c]{90}{Tabular data}}}
& Synth\_low & 5000 & 100 & 10 & 500 & \cite{densired}\\
& Synth\_high & 5000 & 100 & 10 & 500 & \cite{densired}\\
& HAR & 10{,}299 & 561 & 6 & 0 & \cite{uciRepo}\\
& letterrecognition & 20{,}000 & 16 & 26 & 0 & \cite{uciRepo}\\
& htru2 & 17{,}898 & 8 & 2 & 0 & \cite{uciRepo}\\
& Mice & 1{,}077 & 68 & 8 & 0 & \cite{uciRepo}\\
& TCGA & 801 & 20{,}264 & 5 & 0 & \cite{uciRepo}\\
& Pendigits & 10{,}992 & 16 & 10 & 0 & \cite{uciRepo}\\
\midrule
\parbox[t]{2mm}{\multirow{2}{*}{\rotatebox[origin=c]{90}{Video}}}
& Weizmann & 5{,}701 & 77{,}760 & 90 & 0 & \cite{weizmannData}\\
& Keck & 25{,}457 & 120{,}000 & 60 & 0 & \cite{keckData}\\
\midrule
\parbox[t]{2mm}{\multirow{9}{*}{\rotatebox[origin=c]{90}{Image data}}}
& COIL20 & 1{,}440 & 16{,}384 & 20 & 0 & \cite{uciRepo}\\
& COIL100 & 7{,}200 & 49{,}152 & 100 & 0 & \cite{uciRepo}\\
& cmu\_faces & 624 & 960 & 20 & 0 & \cite{uciRepo}\\
& Optdigits & 5{,}620 & 64 & 10 & 0 & \cite{uciRepo}\\
& USPS & 9{,}298 & 256 & 10 & 0 & \cite{usps}\\
& MNIST & 70{,}000 & 784 & 10 & 0 & \cite{mnist}\\
& FMNIST & 70{,}000 & 784 & 10 & 0 & \cite{fmnist}\\
& KMNIST & 70{,}000 & 784 & 10 & 0 & \cite{kmnist}\\
\bottomrule
\end{tabular}}
\end{table}
\begin{table}[!htb]

\centering
\caption{Parameters for the Synth data set}
\label{tab:synthexperiments}
\resizebox{0.85\linewidth}{!}{
\begin{tabular}{l|c|c}
\toprule
\textbf{Parameters} & \multicolumn{2}{c}{\textbf{Values}}\\
\midrule
$n$ & \multicolumn{2}{c}{5000} \\ 
\midrule
dim & 100 & [10, 50, ..., 50000] \\
ratio\_noise & [0.0, 0.1, ..., 0.9] & 0.1 \\ 
\midrule
max\_retry & \multicolumn{2}{c}{5} \\ 
dens\_factors & \multicolumn{2}{c}{[1, 1, 0.5, 0.3, 2, 1.2, 0.9, 0.6, 1.4, 1.1]} \\ 
square & \multicolumn{2}{c}{True} \\ 
clunum & \multicolumn{2}{c}{10} \\ 
seed & \multicolumn{2}{c}{6} \\ 
core\_num & \multicolumn{2}{c}{200} \\ 
momentum & \multicolumn{2}{c}{0.8} \\ 
step & \multicolumn{2}{c}{1.5} \\ 
branch & \multicolumn{2}{c}{0.1} \\ 
star & \multicolumn{2}{c}{1} \\ 
verbose & \multicolumn{2}{c}{False} \\ 
safety & \multicolumn{2}{c}{False} \\ 
domain\_size & \multicolumn{2}{c}{20} \\ 
random\_start & \multicolumn{2}{c}{False} \\
\bottomrule
\end{tabular}}
\end{table}

\newpage

\section{Ablation -- Loss Function}
\label{appendix:loss_extended}
In \Cref{tab:ablation:loss_functions}, the performance of using the individual loss terms and both together are reported for all datasets. This is an extension of \Cref{tab:ablation_loss}.

\begin{table}[bht]
\centering
\caption{Extensive ablation study regarding the loss function, including more datasets and more evaluation measures than in the main part.}
\label{tab:ablation:loss_functions}
\resizebox{1\linewidth}{!}{
\begin{tabular}{l|l|ccc}
\toprule
\textbf{Dataset} & \textbf{Metric} & $\mathcal{L}_D + \mathcal{L}_{rec}$ & $\mathcal{L}_{rec}$ & $\mathcal{L}_D$ \\
\midrule
Synth\_low & ARI & \cellcolor{Green!70}\bm{$99.7 \pm 0.5$} & \cellcolor{Green!5}$86.6 \pm 0.4$ & \cellcolor{Green!57}\underline{$97.1 \pm 3.3$}\\
& NMI & \cellcolor{Green!70}\bm{$99.6 \pm 0.4$} & \cellcolor{Green!5}$90.4 \pm 0.1$ & \cellcolor{Green!60}\underline{$98.2 \pm 1.5$}\\
& DCSI & \cellcolor{Green!20}\underline{$97.5 \pm 0.5$} & \cellcolor{Green!70}\bm{$98.8 \pm 0.2$} & \cellcolor{Green!5}$97.1 \pm 0.5$\\
\midrule
Synth\_high & ARI & \cellcolor{Green!70}\bm{$97.3 \pm 1.5$} & \cellcolor{Green!5}$85.8 \pm 0.2$ & \cellcolor{Green!65}\underline{$96.4 \pm 1.8$}\\
& NMI & \cellcolor{Green!70}\bm{$97.2 \pm 1.4$} & \cellcolor{Green!5}$87.9 \pm 0.0$ & \cellcolor{Green!63}\underline{$96.2 \pm 1.8$}\\
& DCSI & \cellcolor{Green!53}\underline{$95.9 \pm 0.9$} & \cellcolor{Green!70}\bm{$96.8 \pm 1.0$} & \cellcolor{Green!5}$93.4 \pm 1.2$\\
\midrule
HAR & ARI & \cellcolor{Green!70}\bm{$37.8 \pm 7.2$} & \cellcolor{Green!5}$33.0 \pm 4.8$ & \cellcolor{Green!8}\underline{$33.2 \pm 0.0$}\\
& NMI & \cellcolor{Green!70}\bm{$58.8 \pm 5.7$} & \cellcolor{Green!5}$52.1 \pm 2.8$ & \cellcolor{Green!39}\underline{$55.6 \pm 0.1$}\\
& DCSI & \cellcolor{Green!70}\bm{$48.3 \pm 5.2$} & \cellcolor{Green!5}$23.7 \pm 2.5$ & \cellcolor{Green!58}\underline{$43.8 \pm 7.9$}\\
\midrule
letterrec. & ARI & \cellcolor{Green!5}\underline{$23.0 \pm 2.0$} & \cellcolor{Green!70}\bm{$23.3 \pm 1.2$} & \cellcolor{Green!5}\underline{$23.0 \pm 1.2$}\\
& NMI & \cellcolor{Green!54}\underline{$57.6 \pm 0.7$} & \cellcolor{Green!70}\bm{$58.1 \pm 0.6$} & \cellcolor{Green!5}$56.1 \pm 0.6$\\
& DCSI & \cellcolor{Green!70}\bm{$23.0 \pm 1.2$} & \cellcolor{Green!5}$19.4 \pm 1.7$ & \cellcolor{Green!52}\underline{$22.0 \pm 1.6$}\\
\midrule
htru2 & ARI & \cellcolor{Green!37}\underline{$52.4 \pm 33.5$} & \cellcolor{Green!70}\bm{$55.2 \pm 7.3$} & \cellcolor{Green!5}$49.7 \pm 22.9$\\
& NMI & \cellcolor{Green!70}\bm{$39.0 \pm 25.5$} & \cellcolor{Green!65}\underline{$38.6 \pm 4.0$} & \cellcolor{Green!5}$33.4 \pm 15.8$\\
& DCSI & \cellcolor{Green!70}\bm{$0.4 \pm 0.1$} & \cellcolor{Green!27}\underline{$0.2 \pm 0.1$} & \cellcolor{Green!5}$0.1 \pm 0.1$\\
\midrule
Mice & ARI & \cellcolor{Green!70}\bm{$27.9 \pm 4.8$} & \cellcolor{Green!61}\underline{$27.3 \pm 2.5$} & \cellcolor{Green!5}$23.4 \pm 8.4$\\
& NMI & \cellcolor{Green!39}\underline{$47.7 \pm 5.4$} & \cellcolor{Green!70}\bm{$51.2 \pm 1.4$} & \cellcolor{Green!5}$43.9 \pm 14.1$\\
& DCSI & \cellcolor{Green!51}\underline{$24.9 \pm 1.3$} & \cellcolor{Green!70}\bm{$25.5 \pm 1.7$} & \cellcolor{Green!5}$23.4 \pm 1.5$\\
\midrule
TCGA & ARI & \cellcolor{Green!69}\underline{$82.3 \pm 9.9$} & \cellcolor{Green!5}$53.7 \pm 13.4$ & \cellcolor{Green!70}\bm{$82.9 \pm 6.0$}\\
& NMI & \cellcolor{Green!70}\bm{$88.8 \pm 5.3$} & \cellcolor{Green!5}$69.5 \pm 11.6$ & \cellcolor{Green!69}\underline{$88.5 \pm 3.1$}\\
& DCSI & \cellcolor{Green!70}\bm{$77.3 \pm 3.7$} & \cellcolor{Green!5}$60.5 \pm 1.3$ & \cellcolor{Green!26}\underline{$66.0 \pm 3.2$}\\
\midrule
Pendigits & ARI & \cellcolor{Green!46}\underline{$75.6 \pm 0.7$} & \cellcolor{Green!70}\bm{$76.2 \pm 0.6$} & \cellcolor{Green!5}$74.6 \pm 1.2$\\
& NMI & \cellcolor{Green!51}\underline{$83.2 \pm 0.6$} & \cellcolor{Green!70}\bm{$83.6 \pm 0.5$} & \cellcolor{Green!5}$82.2 \pm 0.9$\\
& DCSI & \cellcolor{Green!66}\underline{$52.2 \pm 1.3$} & \cellcolor{Green!5}$33.8 \pm 0.9$ & \cellcolor{Green!70}\bm{$53.4 \pm 1.0$}\\
\midrule
Weizmann & ARI & \cellcolor{Green!25}\underline{$48.7 \pm 2.1$} & \cellcolor{Green!70}\bm{$49.8 \pm 5.6$} & \cellcolor{Green!5}$48.2 \pm 5.0$\\
& NMI & \cellcolor{Green!5}$79.9 \pm 0.9$ & \cellcolor{Green!70}\bm{$81.4 \pm 0.8$} & \cellcolor{Green!22}\underline{$80.3 \pm 1.1$}\\
& DCSI & \cellcolor{Green!5}$72.5 \pm 2.3$ & \cellcolor{Green!70}\bm{$77.0 \pm 2.3$} & \cellcolor{Green!9}\underline{$72.8 \pm 3.6$}\\
\midrule
Keck & ARI & \cellcolor{Green!31}\underline{$7.2 \pm 0.4$} & \cellcolor{Green!5}$7.0 \pm 0.2$ & \cellcolor{Green!70}\bm{$7.5 \pm 0.4$}\\
& NMI & \cellcolor{Green!5}$60.6 \pm 0.5$ & \cellcolor{Green!70}\bm{$62.2 \pm 0.2$} & \cellcolor{Green!25}\underline{$61.1 \pm 0.5$}\\
& DCSI & \cellcolor{Green!61}\underline{$26.9 \pm 2.2$} & \cellcolor{Green!5}$25.6 \pm 1.1$ & \cellcolor{Green!70}\bm{$27.1 \pm 3.3$}\\
\midrule
COIL20 & ARI & \cellcolor{Green!14}\underline{$70.2 \pm 4.7$} & \cellcolor{Green!70}\bm{$74.1 \pm 1.9$} & \cellcolor{Green!5}$69.6 \pm 4.2$\\
& NMI & \cellcolor{Green!22}\underline{$86.0 \pm 2.0$} & \cellcolor{Green!70}\bm{$89.2 \pm 0.4$} & \cellcolor{Green!5}$84.9 \pm 2.4$\\
& DCSI & \cellcolor{Green!26}\underline{$80.6 \pm 3.5$} & \cellcolor{Green!70}\bm{$86.7 \pm 1.7$} & \cellcolor{Green!5}$77.6 \pm 2.1$\\
\midrule
COIL100 & ARI & \cellcolor{Green!9}\underline{$58.0 \pm 3.9$} & \cellcolor{Green!70}\bm{$76.5 \pm 1.4$} & \cellcolor{Green!5}$56.7 \pm 2.4$\\
& NMI & \cellcolor{Green!9}\underline{$85.9 \pm 1.2$} & \cellcolor{Green!70}\bm{$92.0 \pm 0.4$} & \cellcolor{Green!5}$85.5 \pm 0.6$\\
& DCSI & \cellcolor{Green!10}\underline{$79.5 \pm 2.4$} & \cellcolor{Green!70}\bm{$89.6 \pm 0.8$} & \cellcolor{Green!5}$78.7 \pm 2.1$\\
\midrule
cmu\_faces & ARI & \cellcolor{Green!5}$37.4 \pm 4.8$ & \cellcolor{Green!70}\bm{$41.2 \pm 13.3$} & \cellcolor{Green!55}\underline{$40.3 \pm 4.2$}\\
& NMI & \cellcolor{Green!5}$67.6 \pm 3.6$ & \cellcolor{Green!70}\bm{$73.7 \pm 10.5$} & \cellcolor{Green!38}\underline{$70.7 \pm 2.7$}\\
& DCSI & \cellcolor{Green!5}$45.0 \pm 3.0$ & \cellcolor{Green!70}\bm{$56.3 \pm 1.1$} & \cellcolor{Green!26}\underline{$48.6 \pm 2.5$}\\
\bottomrule
\end{tabular}}
\end{table}

\newpage

\section{Real-World Videodata}\label{sec:exp-videos}
We evaluate our approach on two video datasets. 
The first dataset is the Weizmann video dataset \cite{weizmannData}. This dataset consists of $93$ videos showing nine persons performing ten different activities like walking or jumping. This results in $90$ clusters, one for each combination of person and activity. We extract $5{,}701$ $144 \times 180 \times 3$ color frames from the videos and interpret each frame as an individual sample. 
The second dataset is the Keck Gesture video dataset \cite{keckData}. It contains $98$ videos of four persons performing $14$ different gestures. As there are pauses between the gestures, we add `no gesture' as a separate label, resulting in $60$ clusters. We extract $25{,}457$ $480 \times 640 \times 3$ color frames from the videos, which are then downsized to $200 \times 200 \times 3$ due to space limitations. Each frame is again interpreted as an individual sample.

\onecolumn

\section{Extended Table}
\label{appendix:extended_table}
In addition to the ARI scores, as stated in \Cref{tab:experiments_ari}, we evaluate the algorithms using NMI and show the identified number of clusters. These results are given in \Cref{tab:experimentResults}.

\renewcommand{\arraystretch}{1}

\begin{table*}[htb]
\centering
\caption{
Overview of clustering results. As SHADE is the only algorithm handling noise, we report its detected noise ratio in column 1 for clarity (\textcolor{yellow!40!black}{\small{\textit{noise}: $xx.x \pm x.x$}}). 
$k$ gives the number of ground truth clusters in the first column and the number of detected clusters in the other columns (SHADE - DipDECK); otherwise, algorithms have received the ground truth number of classes as an input parameter (-).
We do not include evaluation measures for SHADE in the color coding as these values are computed on labelings including noise, which could lead to an overoptimistic assessment of our method. Note, however, that the clustering quality that SHADE reaches for non-noise points (column 'SHADE') is of exceptionally good quality for most datasets. (*DipEncoder is given the number of clusters, but can `lose' them, leading to a lower $k$)}
\label{tab:experimentResults}
\resizebox{1\textwidth}{!}{
\begin{tabular}{rl|l|c|cccccccc}
\toprule[0.5mm]
& \textbf{Dataset} & \textbf{Metric} & SHADE & SHADE\_1nn & DDC & DipDECK & DipEncoder & DCN & DEC & IDEC\\
\midrule
\midrule
\parbox[t]{2mm}{\multirow{25}{*}{\rotatebox[origin=c]{90}{Tabular data}}}
& Synth\_low & ARI & $99.4 \pm 1.8$ & \cellcolor{Green!70}\bm{$98.9 \pm 2.0$} & \cellcolor{Green!39}\underline{$56.9 \pm 5.5$} & \cellcolor{Green!23}$33.9 \pm 6.4$ & \cellcolor{Green!5}$10.1 \pm 9.9$ & \cellcolor{Green!5}$9.6 \pm 9.7$ & \cellcolor{Green!27}$40.2 \pm 3.0$ & \cellcolor{Green!9}$15.3 \pm 8.8$\\
& \textcolor{yellow!40!black}{\small{(\textit{noise}: $1.1 \pm 1.3$)}} & NMI & $99.7 \pm 1.0$ & \cellcolor{Green!70}\bm{$99.2 \pm 1.2$} & \cellcolor{Green!55}\underline{$82.9 \pm 3.2$} & \cellcolor{Green!32}$58.0 \pm 5.0$ & \cellcolor{Green!7}$31.3 \pm 13.6$ & \cellcolor{Green!5}$29.4 \pm 13.4$ & \cellcolor{Green!42}$69.4 \pm 3.0$ & \cellcolor{Green!19}$44.3 \pm 9.5$\\
& \small{$k=10$} & $k$ & $10.0 \pm 0.0$ & $10.0 \pm 0.0$ & $16.8 \pm 1.0$ & $4.1 \pm 0.5$ & - & - & - & -\\
\cmidrule{2-11}
& Synth\_high & ARI & $98.4 \pm 1.2$ & \cellcolor{Green!70}\bm{$97.5 \pm 1.4$} & \cellcolor{Green!23}\underline{$33.9 \pm 11.1$} & \cellcolor{Green!20}$29.9 \pm 13.6$ & \cellcolor{Green!5}$9.3 \pm 10.7$ & \cellcolor{Green!5}$8.8 \pm 10.5$ & \cellcolor{Green!21}$30.3 \pm 3.5$ & \cellcolor{Green!12}$17.9 \pm 6.6$\\
& \textcolor{yellow!40!black}{\small{(\textit{noise}: $2.4 \pm 1.2$)}} & NMI & $98.1 \pm 1.2$ & \cellcolor{Green!70}\bm{$97.3 \pm 1.3$} & \cellcolor{Green!37}\underline{$61.8 \pm 6.1$} & \cellcolor{Green!29}$53.2 \pm 12.0$ & \cellcolor{Green!7}$28.9 \pm 13.4$ & \cellcolor{Green!5}$26.9 \pm 13.2$ & \cellcolor{Green!37}$61.4 \pm 4.8$ & \cellcolor{Green!23}$46.7 \pm 5.0$\\
& \small{$k=10$} & $k$ & $13.0 \pm 1.5$ & $13.0 \pm 1.5$ & $10.3 \pm 2.1$ & $4.7 \pm 1.3$ & - & - & - & -\\
\cmidrule{2-11}
& HAR & ARI & $36.0 \pm 5.6$ & \cellcolor{Green!5}$36.4 \pm 6.4$ & \cellcolor{Green!33}$49.4 \pm 3.3$ & \cellcolor{Green!38}$51.3 \pm 4.0$ & \cellcolor{Green!57}$60.0 \pm 6.9$ & \cellcolor{Green!70}\bm{$66.1 \pm 1.3$} & \cellcolor{Green!64}$63.4 \pm 2.4$ & \cellcolor{Green!67}\underline{$64.9 \pm 0.9$}\\
& \textcolor{yellow!40!black}{\small{(\textit{noise}: $3.2 \pm 5.2$)}} & NMI & $58.5 \pm 5.8$ & \cellcolor{Green!5}$58.2 \pm 5.5$ & \cellcolor{Green!43}$68.2 \pm 2.1$ & \cellcolor{Green!55}$71.3 \pm 2.0$ & \cellcolor{Green!63}$73.6 \pm 4.1$ & \cellcolor{Green!70}\bm{$75.4 \pm 1.2$} & \cellcolor{Green!68}\underline{$75.0 \pm 1.8$} & \cellcolor{Green!67}$74.6 \pm 0.7$\\
& \small{$k=6$} & $k$ & $3.3 \pm 2.0$ & $3.3 \pm 2.0$ & $4.2 \pm 1.0$ & $3.1 \pm 0.3$ & - & - & - & -\\
\cmidrule{2-11}
& letterrec. & ARI & $43.2 \pm 1.7$ & \cellcolor{Green!62}$23.0 \pm 0.9$ & \cellcolor{Green!14}$9.9 \pm 2.9$ & \cellcolor{Green!5}$7.3 \pm 3.5$ & \cellcolor{Green!68}\underline{$24.7 \pm 1.3$} & \cellcolor{Green!61}$22.6 \pm 1.0$ & \cellcolor{Green!65}$23.9 \pm 1.6$ & \cellcolor{Green!70}\bm{$25.2 \pm 1.8$}\\
& \textcolor{yellow!40!black}{\small{(\textit{noise}: $50.3 \pm 1.8$)}} & NMI & $75.6 \pm 1.0$ & \cellcolor{Green!70}\bm{$57.4 \pm 0.5$} & \cellcolor{Green!31}$43.5 \pm 2.5$ & \cellcolor{Green!5}$34.4 \pm 3.8$ & \cellcolor{Green!48}$49.7 \pm 0.9$ & \cellcolor{Green!39}$46.4 \pm 1.0$ & \cellcolor{Green!51}\underline{$50.8 \pm 1.0$} & \cellcolor{Green!48}$49.7 \pm 1.2$\\
& \small{$k=26$} & $k$ & $111.4 \pm 4.9$ & $111.4 \pm 4.9$ & $14.0 \pm 1.0$ & $13.0 \pm 2.4$ & - & - & - & -\\
\cmidrule{2-11}
& htru2 & ARI & $72.8 \pm 23.4$ & \cellcolor{Green!70}\bm{$65.0 \pm 19.5$} & \cellcolor{Green!54}$49.4 \pm 13.0$ & \cellcolor{Green!12}$9.7 \pm 19.4$ & \cellcolor{Green!6}$4.3 \pm 0.8$ & \cellcolor{Green!54}\underline{$49.7 \pm 2.8$} & \cellcolor{Green!5}$3.0 \pm 0.5$ & \cellcolor{Green!5}$3.2 \pm 0.6$\\
& \textcolor{yellow!40!black}{\small{(\textit{noise}: $14.0 \pm 3.8$)}} & NMI & $59.9 \pm 19.6$ & \cellcolor{Green!70}\bm{$47.4 \pm 14.4$} & \cellcolor{Green!62}\underline{$42.1 \pm 7.0$} & \cellcolor{Green!5}$5.5 \pm 11.0$ & \cellcolor{Green!13}$10.5 \pm 1.4$ & \cellcolor{Green!45}$31.6 \pm 7.0$ & \cellcolor{Green!13}$10.8 \pm 0.6$ & \cellcolor{Green!13}$10.7 \pm 0.6$\\
& \small{$k=2$} & $k$ & $5.9 \pm 2.1$ & $5.9 \pm 2.1$ & $3.7 \pm 0.6$ & $1.2 \pm 0.4$ & - & - & - & -\\
\cmidrule{2-11}
& Mice & ARI & $32.5 \pm 3.8$ & \cellcolor{Green!70}\bm{$27.7 \pm 2.9$} & \cellcolor{Green!43}\underline{$25.2 \pm 1.9$} & \cellcolor{Green!17}$22.7 \pm 4.3$ & \cellcolor{Green!5}$21.6 \pm 2.6$ & \cellcolor{Green!6}$21.7 \pm 1.4$ & \cellcolor{Green!9}$22.0 \pm 1.5$ & \cellcolor{Green!7}$21.8 \pm 1.4$\\
& \textcolor{yellow!40!black}{\small{(\textit{noise}: $30.0 \pm 5.5$)}} & NMI & $56.2 \pm 4.6$ & \cellcolor{Green!65}\underline{$48.7 \pm 2.3$} & \cellcolor{Green!70}\bm{$49.6 \pm 2.0$} & \cellcolor{Green!61}$48.0 \pm 5.3$ & \cellcolor{Green!7}$38.7 \pm 2.6$ & \cellcolor{Green!5}$38.3 \pm 1.5$ & \cellcolor{Green!11}$39.4 \pm 1.8$ & \cellcolor{Green!5}$38.3 \pm 1.8$\\
& \small{$k=8$} & $k$ & $11.9 \pm 2.8$ & $11.9 \pm 2.8$ & $15.3 \pm 1.2$ & $13.4 \pm 5.3$ & - & - & - & -\\
\cmidrule{2-11}
& TCGA & ARI & $86.6 \pm 7.8$ & \cellcolor{Green!5}$80.0 \pm 13.7$ & \cellcolor{Green!41}$87.5 \pm 0.8$ & \cellcolor{Green!48}\underline{$88.8 \pm 4.4$} & \cellcolor{Green!70}\bm{$93.4 \pm 6.0$} & \cellcolor{Green!40}$87.2 \pm 5.3$ & \cellcolor{Green!30}$85.1 \pm 2.7$ & \cellcolor{Green!18}$82.6 \pm 0.9$\\
& \textcolor{yellow!40!black}{\small{(\textit{noise}: $21.8 \pm 9.0$)}} & NMI & $90.8 \pm 4.4$ & \cellcolor{Green!16}$87.4 \pm 7.2$ & \cellcolor{Green!53}\underline{$91.9 \pm 0.7$} & \cellcolor{Green!53}\underline{$91.9 \pm 2.1$} & \cellcolor{Green!70}\bm{$94.0 \pm 3.8$} & \cellcolor{Green!30}$89.1 \pm 3.1$ & \cellcolor{Green!33}$89.5 \pm 1.2$ & \cellcolor{Green!5}$86.1 \pm 1.4$\\
& \small{$k=5$} & $k$ & $5.7 \pm 0.9$ & $5.7 \pm 0.9$ & $6.0 \pm 0.0$ & $5.9 \pm 0.5$ & - & - & - & -\\
\cmidrule{2-11}
& Pendigits & ARI & $85.1 \pm 1.4$ & \cellcolor{Green!62}\underline{$75.1 \pm 0.8$} & \cellcolor{Green!70}\bm{$76.9 \pm 2.0$} & \cellcolor{Green!59}$74.3 \pm 1.1$ & \cellcolor{Green!18}$64.6 \pm 3.0$ & \cellcolor{Green!5}$61.6 \pm 1.9$ & \cellcolor{Green!22}$65.7 \pm 3.3$ & \cellcolor{Green!19}$64.9 \pm 2.6$\\
& \textcolor{yellow!40!black}{\small{(\textit{noise}: $17.5 \pm 2.2$)}} & NMI & $89.3 \pm 0.9$ & \cellcolor{Green!62}\underline{$82.7 \pm 0.6$} & \cellcolor{Green!70}\bm{$84.1 \pm 1.0$} & \cellcolor{Green!58}$82.0 \pm 0.8$ & \cellcolor{Green!19}$75.2 \pm 1.3$ & \cellcolor{Green!5}$72.7 \pm 0.6$ & \cellcolor{Green!28}$76.7 \pm 1.4$ & \cellcolor{Green!23}$75.9 \pm 0.9$\\
& \small{$k=10$} & $k$ & $20.0 \pm 1.8$ & $20.0 \pm 1.8$ & $13.0 \pm 0.4$ & $14.8 \pm 0.7$ & - & - & - & -\\
\midrule
\midrule
\parbox[t]{2mm}{\multirow{6}{*}{\rotatebox[origin=c]{90}{Video data}}}
& Weizmann & ARI & $57.1 \pm 5.9$ & \cellcolor{Green!70}\bm{$48.2 \pm 3.6$} & \cellcolor{Green!10}$14.7 \pm 1.8$ & \cellcolor{Green!5}$12.0 \pm 1.9$ & \cellcolor{Green!25}$23.3 \pm 1.2$ & \cellcolor{Green!28}$24.6 \pm 1.1$ & \cellcolor{Green!28}\underline{$24.9 \pm 1.2$} & \cellcolor{Green!28}$24.7 \pm 1.2$\\
& \textcolor{yellow!40!black}{\small{(\textit{noise}: $14.6 \pm 2.6$)}} & NMI & $86.6 \pm 0.6$ & \cellcolor{Green!70}\bm{$80.2 \pm 1.1$} & \cellcolor{Green!16}$57.5 \pm 2.0$ & \cellcolor{Green!5}$52.8 \pm 1.9$ & \cellcolor{Green!27}$61.9 \pm 0.8$ & \cellcolor{Green!28}$62.3 \pm 0.9$ & \cellcolor{Green!31}\underline{$63.7 \pm 1.0$} & \cellcolor{Green!30}$63.3 \pm 1.0$\\
& \small{$k=90$} & $k$ & $57.1 \pm 2.1$ & $57.1 \pm 2.1$ & $20.8 \pm 2.4$ & $24.5 \pm 2.2$ & $88.4 \pm 1.2^*$ & - & - & -\\
\cmidrule{2-11}
& Keck & ARI & $9.3 \pm 0.5$ & \cellcolor{Green!70}\bm{$7.5 \pm 0.4$} & \cellcolor{Green!5}$-0.2 \pm 1.1$ & \cellcolor{Green!65}$6.9 \pm 0.8$ & \cellcolor{Green!67}\underline{$7.1 \pm 0.3$} & \cellcolor{Green!61}$6.4 \pm 0.5$ & \cellcolor{Green!58}$6.1 \pm 0.9$ & \cellcolor{Green!59}$6.2 \pm 0.9$\\
& \textcolor{yellow!40!black}{\small{(\textit{noise}: $25.4 \pm 1.1$)}} & NMI & $66.3 \pm 0.3$ & \cellcolor{Green!70}\bm{$61.6 \pm 0.3$} & \cellcolor{Green!5}$18.1 \pm 4.3$ & \cellcolor{Green!30}$34.9 \pm 6.0$ & \cellcolor{Green!41}\underline{$42.4 \pm 0.6$} & \cellcolor{Green!37}$39.4 \pm 2.5$ & \cellcolor{Green!38}$39.9 \pm 3.5$ & \cellcolor{Green!37}$39.2 \pm 5.2$\\
& \small{$k=60$} & $k$ & $226.6 \pm 10.4$ & $226.6 \pm 10.4$ & $10.6 \pm 3.1$ & $32.3 \pm 7.4$ & - & - & - & -\\
\midrule
\midrule
\parbox[t]{2mm}{\multirow{9}{*}{\rotatebox[origin=c]{90}{Image data}}}
& COIL20 & ARI & $82.5 \pm 4.5$ & \cellcolor{Green!70}\bm{$68.7 \pm 3.5$} & \cellcolor{Green!46}$62.0 \pm 5.5$ & \cellcolor{Green!5}$50.5 \pm 7.8$ & \cellcolor{Green!53}\underline{$64.0 \pm 3.0$} & \cellcolor{Green!47}$62.4 \pm 2.8$ & \cellcolor{Green!52}$63.7 \pm 2.8$ & \cellcolor{Green!49}$62.9 \pm 2.9$\\
& \textcolor{yellow!40!black}{\small{(\textit{noise}: $12.5 \pm 1.9$)}} & NMI & $93.6 \pm 1.7$ & \cellcolor{Green!70}\bm{$85.6 \pm 1.3$} & \cellcolor{Green!69}\underline{$85.5 \pm 0.9$} & \cellcolor{Green!8}$79.9 \pm 2.4$ & \cellcolor{Green!12}$80.2 \pm 1.1$ & \cellcolor{Green!5}$79.6 \pm 1.3$ & \cellcolor{Green!16}$80.6 \pm 1.0$ & \cellcolor{Green!9}$80.0 \pm 1.1$\\
& \small{$k=20$} & $k$ & $16.5 \pm 1.0$ & $16.5 \pm 1.0$ & $14.3 \pm 0.8$ & $18.9 \pm 1.0$ & - & - & - & -\\
\cmidrule{2-11}
& COIL100 & ARI & $78.1 \pm 7.3$ & \cellcolor{Green!70}\underline{$56.8 \pm 5.0$} & \cellcolor{Green!5}$16.4 \pm 3.8$ & \cellcolor{Green!13}$21.4 \pm 3.0$ & \cellcolor{Green!66}$54.3 \pm 1.9$ & \cellcolor{Green!68}$55.9 \pm 3.0$ & \cellcolor{Green!68}$55.8 \pm 2.0$ & \cellcolor{Green!70}\bm{$56.9 \pm 2.0$}\\
& \textcolor{yellow!40!black}{\small{(\textit{noise}: $24.9 \pm 1.7$)}} & NMI & $94.4 \pm 0.9$ & \cellcolor{Green!68}\underline{$85.4 \pm 0.6$} & \cellcolor{Green!5}$69.5 \pm 3.0$ & \cellcolor{Green!5}$69.5 \pm 1.7$ & \cellcolor{Green!57}$82.6 \pm 0.6$ & \cellcolor{Green!64}$84.2 \pm 0.7$ & \cellcolor{Green!70}\bm{$85.8 \pm 0.5$} & \cellcolor{Green!66}$84.7 \pm 0.5$\\
& \small{$k=100$} & $k$ & $77.8 \pm 3.2$ & $77.8 \pm 3.2$ & $18.3 \pm 3.3$ & $26.8 \pm 1.9$ & $99.8 \pm 0.4$ & - & - & -\\
\cmidrule{2-11}
& cmu\_faces & ARI & $38.9 \pm 7.6$ & \cellcolor{Green!35}$34.6 \pm 6.2$ & \cellcolor{Green!37}$35.0 \pm 3.5$ & \cellcolor{Green!5}$29.8 \pm 9.8$ & \cellcolor{Green!55}$37.9 \pm 2.2$ & \cellcolor{Green!70}\bm{$40.3 \pm 2.0$} & \cellcolor{Green!42}$35.8 \pm 2.8$ & \cellcolor{Green!64}\underline{$39.4 \pm 3.3$}\\
& \textcolor{yellow!40!black}{\small{(\textit{noise}: $15.1 \pm 4.1$)}} & NMI & $69.1 \pm 5.3$ & \cellcolor{Green!33}$65.0 \pm 4.9$ & \cellcolor{Green!27}$64.4 \pm 2.3$ & \cellcolor{Green!5}$62.3 \pm 7.6$ & \cellcolor{Green!62}$67.7 \pm 1.6$ & \cellcolor{Green!70}\bm{$68.5 \pm 1.0$} & \cellcolor{Green!45}$66.1 \pm 2.0$ & \cellcolor{Green!67}\underline{$68.2 \pm 2.2$}\\
& \small{$k=20$} & $k$ & $10.6 \pm 1.7$ & $10.6 \pm 1.7$ & $12.0 \pm 0.9$ & $8.9 \pm 2.2$ & - & - & - & -\\
\bottomrule
\bottomrule
\end{tabular}}
\end{table*}

\renewcommand{\arraystretch}{1.0}

\newpage

\section{Benchmark Datasets with Gaussian Clusters}
\label{appendix:mnist}
In \Cref{tab:mnist_etc}, we include results on the most common benchmark datasets in the area: Optdigits, Pendigits, USPS, MNIST, FMNIST, and KMNIST.
Note that they have in common that their classes are centroid-based: each object of a class is similar to a concept given by the label, e.g., the picture of a written digit. This results in a typical, roughly Gaussian-like distribution where most points of a class are rather similar to the center, and fewer points are far away from the center. While SHADE returns competitive NMI and ARI values on the clustered data points (column 'SHADE'), it detects objects far away from their concept as noise, which can be the majority of the data in the worst case.

\begin{table*}[htb]

\centering
\caption{Analogous to \Cref{tab:experimentResults}, we show further clustering results for data sets that do not contain density-based structures, but are, however, often expected as benchmarks for deep clustering. SHADE detects high percentages of these data sets as noise as especially points at the border of Gaussian clusters have often no density-connection to the rest of the cluster.}
\label{tab:mnist_etc} 
\resizebox{1\textwidth}{!}{
\begin{tabular}{l|l|c|ccccccc}
\toprule
\textbf{Dataset} & \textbf{Metric} & SHADE & SHADE\_1nn & DDC & DipDECK & DipEncoder & DCN & DEC & IDEC\\
\midrule
Optdigits & ARI & $93.3 \pm 1.9$ & \cellcolor{Green!10}$78.0 \pm 7.4$ & \cellcolor{Green!70}\bm{$88.9 \pm 2.3$} & \cellcolor{Green!33}\underline{$82.2 \pm 2.3$} & \cellcolor{Green!22}$80.2 \pm 4.0$ & \cellcolor{Green!5}$77.0 \pm 3.5$ & \cellcolor{Green!24}$80.4 \pm 3.1$ & \cellcolor{Green!25}$80.7 \pm 3.6$\\
\textcolor{yellow!40!black}{\small{(\textit{noise}: $38.2 \pm 5.3$)}} & NMI & $94.6 \pm 1.0$ & \cellcolor{Green!9}$83.4 \pm 3.9$ & \cellcolor{Green!70}\bm{$91.7 \pm 1.3$} & \cellcolor{Green!29}$86.2 \pm 0.9$ & \cellcolor{Green!25}$85.6 \pm 2.0$ & \cellcolor{Green!5}$82.9 \pm 1.6$ & \cellcolor{Green!30}\underline{$86.3 \pm 1.4$} & \cellcolor{Green!29}$86.2 \pm 1.5$\\
\small{$k=10$} & $k$ & $12.4 \pm 1.0$ & $12.4 \pm 1.0$ & $11.1 \pm 0.5$ & $11.0 \pm 1.2$ & - & - & - & -\\
\midrule
USPS & ARI & $88.1 \pm 2.7$ & \cellcolor{Green!10}$68.1 \pm 1.5$ & \cellcolor{Green!70}\bm{$92.2 \pm 2.2$} & \cellcolor{Green!10}$68.2 \pm 5.1$ & \cellcolor{Green!21}$72.5 \pm 2.0$ & \cellcolor{Green!5}$66.2 \pm 1.3$ & \cellcolor{Green!23}$73.4 \pm 0.9$ & \cellcolor{Green!24}\underline{$74.0 \pm 0.9$}\\
\textcolor{yellow!40!black}{\small{(\textit{noise}: $45.7 \pm 2.8$)}} & NMI & $91.8 \pm 1.1$ & \cellcolor{Green!11}$75.9 \pm 1.1$ & \cellcolor{Green!70}\bm{$91.0 \pm 0.8$} & \cellcolor{Green!21}$78.5 \pm 2.5$ & \cellcolor{Green!26}$79.9 \pm 1.6$ & \cellcolor{Green!5}$74.5 \pm 1.0$ & \cellcolor{Green!31}$81.0 \pm 0.5$ & \cellcolor{Green!32}\underline{$81.3 \pm 0.6$}\\
\small{$k=10$} & $k$ & $9.5 \pm 0.9$ & $9.5 \pm 0.9$ & $9.8 \pm 0.4$ & $8.0 \pm 1.2$ & - & - & - & -\\
\midrule
MNIST & ARI & $86.1 \pm 8.4$ & \cellcolor{Green!5}$54.1 \pm 2.8$ & \cellcolor{Green!70}\bm{$94.5 \pm 1.2$} & \cellcolor{Green!48}$80.9 \pm 3.0$ & \cellcolor{Green!46}$79.6 \pm 2.8$ & \cellcolor{Green!50}\underline{$82.0 \pm 4.5$} & \cellcolor{Green!46}$79.5 \pm 2.1$ & \cellcolor{Green!50}$81.9 \pm 2.0$\\
\textcolor{yellow!40!black}{\small{(\textit{noise}: $58.9 \pm 5.3$)}} & NMI & $84.5 \pm 2.4$ & \cellcolor{Green!5}$63.7 \pm 1.3$ & \cellcolor{Green!70}\bm{$93.7 \pm 0.8$} & \cellcolor{Green!50}$84.5 \pm 1.5$ & \cellcolor{Green!51}$84.9 \pm 1.6$ & \cellcolor{Green!50}$84.6 \pm 1.9$ & \cellcolor{Green!51}$85.1 \pm 1.0$ & \cellcolor{Green!57}\underline{$87.5 \pm 0.8$}\\
\small{$k=10$} & $k$ & $38.0 \pm 9.9$ & $38.0 \pm 9.9$ & $10.0 \pm 0.0$ & $11.3 \pm 0.8$ & - & - & - & -\\
\midrule
FMNIST & ARI & $72.2 \pm 2.7$ & \cellcolor{Green!6}$35.7 \pm 1.4$ & \cellcolor{Green!5}$35.5 \pm 7.1$ & \cellcolor{Green!68}\underline{$46.5 \pm 0.9$} & \cellcolor{Green!51}$43.5 \pm 3.6$ & \cellcolor{Green!64}$45.8 \pm 2.9$ & \cellcolor{Green!44}$42.3 \pm 4.1$ & \cellcolor{Green!70}\bm{$46.9 \pm 3.6$}\\
\textcolor{yellow!40!black}{\small{(\textit{noise}: $66.3 \pm 2.3$)}} & NMI & $73.4 \pm 1.0$ & \cellcolor{Green!5}$53.9 \pm 0.9$ & \cellcolor{Green!63}\underline{$64.1 \pm 3.3$} & \cellcolor{Green!70}\bm{$65.3 \pm 0.7$} & \cellcolor{Green!36}$59.4 \pm 2.4$ & \cellcolor{Green!53}$62.4 \pm 2.2$ & \cellcolor{Green!36}$59.3 \pm 2.7$ & \cellcolor{Green!61}$63.7 \pm 2.5$\\
\small{$k=10$} & $k$ & $68.7 \pm 9.2$ & $68.7 \pm 9.2$ & $7.1 \pm 0.9$ & $9.6 \pm 1.1$ & - & - & - & -\\
\midrule
KMNIST & ARI & $58.6 \pm 9.4$ & \cellcolor{Green!5}$27.5 \pm 1.7$ & \cellcolor{Green!70}\bm{$60.3 \pm 3.5$} & \cellcolor{Green!17}$33.8 \pm 9.9$ & \cellcolor{Green!37}\underline{$43.6 \pm 1.1$} & \cellcolor{Green!30}$40.3 \pm 1.6$ & \cellcolor{Green!35}$42.4 \pm 0.9$ & \cellcolor{Green!35}$42.8 \pm 1.1$\\
\textcolor{yellow!40!black}{\small{(\textit{noise}: $73.4 \pm 2.2$)}} & NMI & $66.2 \pm 3.1$ & \cellcolor{Green!5}$46.8 \pm 0.6$ & \cellcolor{Green!70}\bm{$73.6 \pm 1.6$} & \cellcolor{Green!27}$55.9 \pm 5.0$ & \cellcolor{Green!29}\underline{$56.7 \pm 1.3$} & \cellcolor{Green!21}$53.6 \pm 1.4$ & \cellcolor{Green!27}$55.7 \pm 1.1$ & \cellcolor{Green!29}\underline{$56.7 \pm 1.2$}\\
\small{$k=10$} & $k$ & $64.1 \pm 10.4$ & $64.1 \pm 10.4$ & $16.2 \pm 0.7$ & $11.7 \pm 2.8$ & - & - & - & -\\
\bottomrule
\end{tabular}}
\end{table*}

\end{document}